\documentclass[twoside,leqno,twocolumn]{article}
\usepackage{ltexpprt}
\usepackage[numbers,sort]{natbib}
\usepackage{amsmath,amssymb}
\usepackage[nocompress]{cite}
\usepackage{algorithm}
\usepackage{algorithmic}
\usepackage{caption}
\usepackage{graphicx}
\usepackage{ltablex}
\usepackage{booktabs}
\usepackage{hyperref}
\usepackage{subcaption}
\usepackage{url}
\usepackage{float}
\usepackage{multicol}
\usepackage{array}
\usepackage{mathrsfs}
\usepackage{color}
\usepackage{breqn}
\usepackage{xtab}
\usepackage{atbegshi}
\AtBeginDocument{\AtBeginShipoutNext{\AtBeginShipoutDiscard}}

\newcolumntype{L}[1]{>{\raggedright\let\newline\\\arraybackslash\hspace{0pt}}m{#1}}
\newcolumntype{C}[1]{>{\centering\let\newline\\\arraybackslash\hspace{0pt}}m{#1}}
\newcolumntype{R}[1]{>{\raggedleft\let\newline\\\arraybackslash\hspace{0pt}}m{#1}}
\newcommand{\bx}{\ensuremath{\mathbf{x}}}
\newcommand{\bX}{\ensuremath{\mathbf{X}}}
\newcommand{\bz}{\ensuremath{\mathbf{z}}}
\newcommand{\bZ}{\ensuremath{\mathbf{Z}}}
\newcommand{\bw}{\ensuremath{\mathbf{w}}}

\newcommand{\bb}{\ensuremath{\mathbf{b}}}
\DeclareMathOperator*{\argmin}{arg\,min}

\newcommand*{\mathcolor}{}
\def\mathcolor#1#{\mathcoloraux{#1}}
\newcommand*{\mathcoloraux}[3]{%
	\protect\leavevmode
	\begingroup
	\color#1{#2}#3%
	\endgroup
}

\begin{document}

\twocolumn[\title{\Large Generalized Inverse Classification}

\author{Michael T. Lash\thanks{Department of Computer Science, University of Iowa.}\\
	\and
	Qihang Lin\thanks{Department of Management Science, University of Iowa.}\\
	\and
	Nick Street\footnotemark[2]\\
	\and
	Jennifer G. Robinson\thanks{Department of Epidemiology, University of Iowa.  \newline \{michael-lash, qihang-lin, nick-street, jennifer-g-robinson, jeffrey-ohlmann\}@uiowa.edu}\\
	\and
	Jeffrey Ohlmann\footnotemark[2]}

\date{}]


\maketitle

\begin{abstract} \small
Inverse classification is the process of perturbing an instance in a meaningful way such that it is more likely to conform to a specific class. Historical methods that address such a problem are often framed to leverage only a single classifier, or specific set of classifiers. These works are often accompanied by naive assumptions. In this work we propose generalized inverse classification (GIC), which avoids restricting the classification model that can be used. We incorporate this formulation into a refined framework in which GIC takes place. Under this framework, GIC operates on features that are immediately actionable. Each change incurs an individual cost, either linear or non-linear. Such changes are subjected to occur within a specified level of cumulative change (budget). Furthermore, our framework incorporates the estimation of features that change as a consequence of direct actions taken (indirectly changeable features). To solve such a problem, we propose three real-valued heuristic-based methods and two sensitivity analysis-based comparison methods, each of which is evaluated on two freely available real-world datasets. Our results demonstrate the validity and benefits of our formulation, framework, and methods. 

\end{abstract}

\section{Introduction}

In typical classification settings, a model is trained and used to make predictions about some event of interest. Depending upon the predictive task, some action may then be taken. In a medical domain, a patient may be monitored more carefully if a prediction yields a high likelihood of some negative outcome. However, in the same setting, we may want to know what actions can be taken to minimize the patient's chances of said adverse event occurring. The process of finding the optimal set of actions, or changes, that can be taken in order to minimize the probability of such events occurring is what we term \textit{inverse classification}.

This example domain further highlights the nature and importance of the problem. Consider, specifically, the problem of mitigating the long-term risk of cardiovascular disease (CVD) of Patient 29 taken from our experiments below. Initially, we use a constructed model to estimate this patient's risk, or probability, of developing CVD, which is found to be 55\%. This estimate is based on pertinent factors such as medications, lab measurements (e.g.,~ blood glucose), lifestyle (e.g.,~diet), and demographics (e.g.,~age).

Following an initial assessment of risk, we would like to work `backwards' through our learned model to obtain recommendations that reduce Patient 29's probabililty of CVD. Past methods, however, restrict the set of classifiers that are used to obtain such recommendations, often only affording the use of a single algorithm. Such restrictions are prohibitive in that a particular classifier may have useful properties. These might include high predictive accuracy, such as the random forest used to obtain Patient 29's initial level of risk, or a high degree of explanatory power,which may help the patient better understand why certain recommendations were made. Therefore we propose \textit{generalized inverse classification} (GIC), which permits the use of virtually any classification function, requiring only a simple non-prohibitive assumption (further discussed in Section 3). This is the first contribution of this work.

Our second contribution is to show that the problem can be solved using heuristics. Specifically, we propose three real-valued heuristic-based methods that solve this problem, which we compare to two sensitivity analysis-based baseline methods. \textcolor{black}{We demonstrate the efficacy of these results on two freely available datasets, one of which includes Patient 29, whose risk we lower from 55\% to less than 30\% (Section 4)}. Thirdly, we refine an existing inverse classification framework to include non-linear cost-to-change functions, which we then incorporate into our experiments. Section 3 outlines the framework, the \textit{generalized inverse classification problem}, the three heuristic-based methods, and two sensitivity analysis-based methods, while Section 5 concludes the paper.

\section{Related Work}
Inverse classification is akin to the sub-discipline of sensitivity analysis, which examines the impact of predictive algorithm input on the output. While there are many forms of sensitivity analysis\citep{isukapalli1999,Yao2003}, local and variable perturbation methods are most similar. Based on this we develop two sensitivity analysis-based methods, related in Section 3, for comparison purposes.


Past works on inverse classification differ with respect to three distinct perspectives: operational data types, algorithmic mechanism, and framework. The \textbf{operational data types} which encode the data, on which inverse classification is performed, are either discrete \citep{Aggarwal2010,Chi2012,Yang2012}, continuous \citep{Mannino2000,Barbella2009,Pendharkar2002,Lash2016}, or both \citep{Lash2016early}. The latter two allow for more fine-grained results leading to greater precision in the recommendations made. 

The \textbf{algorithmic mechanism} operates on these data types by finding the feasible recommendations 
that optimize the predicted probability. Such optimization strategies are constructed to be greedy \citep{Aggarwal2010,Chi2012,Yang2012,Mannino2000} or non-greedy \citep{Barbella2009,Pendharkar2002,Lash2016,Lash2016early}.


The \textbf{framework} ensures that recommended changes are feasible and implementable. These include: (1) identifying features that can be changed (e.g.~age cannot), (2) the difficulty in implementing changes (feature-specific costs) and (3) a restriction on the cumulative change (budget). In \citep{Aggarwal2010,Pendharkar2002} there are no constraints imposed. Of those that do impose constrains:
\begin{itemize}
	\setlength\itemsep{0em}
	\item In \citep{Barbella2009} constraints are imposed that lead to non-extreme recommendations, but neither (1), (2), or (3) are considered.
	\item In \citep{Mannino2000} (2) is imposed, but not (1) or (3).
	\item In \citep{Chi2012,Yang2012} only (1) and (2) are considered.
	\item A different notion of (1), (2), and (3) are explored in \citep{Lash2016early} by matching discrete entities to compute features.
	\item In \citep{Lash2016} (1), (2), and (3) are all considered, but does not permit nondifferentiable classifiers.
\end{itemize}

Real-valued heuristic-based methods are also relevant to this work. These methods include variable neighborhood search (VNS), genetic algorithms \citep{Michalewicz2013}, and hill-climbing \citep{Chi2012,Mannino2000}. In this work we elect to focus on genetic algorithms, hill-climbing, and local search, which can be viewed as a simpler form of VNS. As will be shown, by using heuristic-based methods, we can be as general as possible in solving the inverse classification problem.




\section{Generalized Inverse Classification}

In this section we first briefly discuss GIC. Subsequently, we outline our inverse classification framework. Next, we relate three heuristic-based methods that can be used to solve GIC. Finally, we introduce two sensitivity analysis-based methods that will be compared to our heuristic-based methods. 

Under the GIC formulation no assumptions are made about the classification function $f(\dot)$ other than $f:\mathbb{R}^{p}\rightarrow \mathbb{R}$. Such a level of generality allow us to obtain optimal solutions for nondifferentiable functions. These functions include popular ensemble techniques such as bagging \citep{Breiman1996} and boosting \citep{Freund1996}, as well as C4.5 decision trees. Classifiers such as these are often found to have high predictive power (ensembles) or are more readily interpretable and explainable (e.g.,~ C4.5 decision trees), which is why it is so important methods be developed that incorporate such classifiers.

\subsection{Framework}
Suppose $\{(\bx^i,y^i)\}_{i=1,2,\dots,n}$ is a dataset of $n$ instances where $\bx^i\in\mathbb{R}^{p}$ is a column feature vector of length $p$ and $y^i \in \{-1,1\}$ is the binary label associated with $\bx^i$ for $i=1,2,\dots,n$. Let $f(\bx)$ be a function that computes the probability of $\bx$ being in the positive class (with $y=1$). Typically, $f(\bx)$ is based on a certain classification model built on the dataset. Given a new instance, with feature vector $\bar\bx$, we want to modify some components of $\bar\bx$, subject to some budget constraints, so that the predicted probability of being positive is minimized.

We further partition the features into three subsets, $U$, $D$ and $I$, which represent the sets of unchangeable, directly changeable and indirectly changeable features, respectively. When we optimize the features, we can only determine the value for $\bx_D$ and the values of $\bx_I$ will depend on $\bx_D$ and $\bx_U$. Therefore, we model the dependency of $\bx_I$ on $\bx_D$ and $\bx_U$ as $\bx_I=H(\bx_D,\bx_U)$ where the mapping $H:\mathbb{R}^{|D|+|U|}\rightarrow\mathbb{R}^{|I|}$ is assumed to be differentiable. Note that the mapping $H$ can be any predictive model constructed using the same training instances. Therefore, we represent $f(\bx)$ as $f(\bx_U,\bx_I,\bx_D)$ to distinguish these three blocks so that the feature optimization problem can be formulated as
\begin{align}
\label{FeatureOptGen}
\min_{\bx_D\in\mathbb{R}^{|D|}}&&f(\bar\bx_U,\bx_I,\bx_D)=f(\bar\bx_U,\cdots \\[-10pt]\nonumber  & &  H(\bx_{D},\bar\bx_U),\bx_D)\\\nonumber
\text{s.t.}&&\phi(\bx_D-\bar\bx_D)\leq B\\\nonumber
&&l_i\leq x_i\leq u_i\text{ for }i\in D.
\end{align}
Here, we assume the reasonable value of each directly changeable feature in $D$ must be within an interval, denoted by $[l_i,u_i]$ for $i\in D$. If $x_i$ can only be increased (decreased), we can set $l_i=\bar x_i$ ($u_i=\bar x_i$).
In addition, $\phi:\mathbb{R}^{|D|}\rightarrow\mathbb{R}$ is a convex cost function that measures the cost for changing $\bar\bx_D$ to $\bx_D$ and $B$ is the total budget we have to support this change. We require $\phi(\mathbf{0})=0$.

Here, we provide two examples of  $\phi(\bz)$. The first assumes the cost increases linearly, as $\bx$ is deviated from $\bar\bx$, which is
\begin{eqnarray}
\label{FeasibleSetLinear}
\phi(\bz)=\sum_{i\in D}c_i^+(z_i)_++c_i^-(z_i)_-
\end{eqnarray}
where $(z)_+=\max\{0,z\}$ and $(z)_-=\max\{0,-z\}$, and $c_i^+$ and $c_i^-$ denotes the costs for increasing and decreasing the feature $x_i$ by one unit for $i\in D$. If one assumes the costs increase quadratically as $\bx$ deviates from $\bar\bx$, then
\begin{eqnarray}
\label{FeasibleSetQuadratic}
\phi(\bz)=\sum_{i\in D}c_i^+(z_i)_+^2+c_i^-(z_i)_-^2
\end{eqnarray}
Note that the constants $c_i^+$ and $c_i^-$ in \eqref{FeasibleSetLinear} and \eqref{FeasibleSetQuadratic} can be different. In both cost functions, if decreasing (increasing) $x_i$ is cost-free, we can set $c_i^-=0$ ($c_i^+=0$). In the rest of this paper, we will only focus on the quadratic cost $\phi(\bz)$ in \eqref{FeasibleSetQuadratic}.

We define $\bz=\bx_D-\bar\bx_D$ in \eqref{FeatureOptGen} and, by changing variables, \eqref{FeatureOptGen} can be equivalently written as
\begin{eqnarray}
\label{FeatureOptGenReform}
\min_{\bz\in\Delta_{D}}&&g(\bz)
\end{eqnarray}
where $g(\bz)\equiv f(\bar\bx_U,H(\bar\bx_D+\bz,\bar\bx_U),\bar\bx_D+\bz)$,
\begin{align}
\label{FeasibleSet}
\Delta_{D}\equiv\left\{\bz\in\mathbb{R}^{|D|}\bigg|
\begin{array}{c}
\phi(\bz)\leq B,\\
l_i'\leq z_i\leq u_i'\text{ for }i\in D.
\end{array}
\right\},
\end{align}
$l_i'=l_i-\bar x_i$ and $u_i'=u_i-\bar x_i$ for $i\in D$. The projection mapping onto the feasible set $\Delta_{D}$ is defined as
\begin{eqnarray}
\label{proj}
\textbf{Proj}_{\Delta_{D}}(\bw)\equiv\argmin_{\bz\in\Delta_{D}}\frac{1}{2}\|\bz-\bw\|^2.
\end{eqnarray}

We then define a subroutine for solving \eqref{proj}. We first define
\begin{align}
\label{h}
h_i(w,\lambda)=
\left\{
\begin{array}{rl}
\max\{\min\{w/(1+2\lambda c_i^+),&\text{ if } w\geq 0 \\\dots u_i'\},l_i'\}\\
\max\{\min\{w/(1+2\lambda c_i^-),&\text{ if } w< 0\\\dots u_i'\},l_i'\}\\
\end{array}
\right.
\end{align}
for each $i\in D$ and $\lambda\geq0$. The subroutine is given in Algorithm \ref{algo:Proj} whose validity can be easily verified by the KKT conditions of \eqref{FeatureOptGenReform}. Note that the bisection search in Algorithm \ref{algo:Proj} can always succeed because $\sum_{i\in D}c_i^+(h_i(w_i,\lambda))_+^2 +c_i^-(h_i(w_i,\lambda)))_-^2$ monotonically decreases to zero as $\lambda$ increases to infinity.
\begin{algorithm}
	\caption{Projection Mapping $\textbf{Proj}$}
	\label{algo:Proj}
	\begin{algorithmic}[1]
        \REQUIRE $\bw\in\mathbb{R}^{|D|}$, $\{c_i^+\}_{i\in D}$, $\{c_i^-\}_{i\in D}$, $\{l'_i\}_{i\in D}$ and $\{u'_i\}_{i\in D}$
        \IF {\small$\sum_{i\in D}c_i^+(h_i(w_i,0))_+^2 +c_i^-(h_i(w_i,0)))_-^2\leq B$\normalsize}
        \STATE{$\lambda\leftarrow 0$}
        \ELSE
        \STATE{Apply bisection search to find $\lambda\in(0,+\infty)$ such that
        	\small
        	\begin{eqnarray*}
        		\sum_{i\in D}c_i^+(h_i(w_i,\lambda))_+^2 +c_i^-(h_i(w_i,\lambda)))_-^2= B
        	\end{eqnarray*}
        	\normalsize}
        \ENDIF
        \STATE {$z_i\leftarrow h_i(w,\lambda)$ for $i\in D$.}
        \ENSURE $\bz$
	\end{algorithmic}
\end{algorithm}

\subsection{Heuristic-based methods}

We propose three real-valued heuristic-based algorithms to solve the generalized inverse classification problem: hill-climbing + local search (HC+LS), a genetic algorithm (GA), and a genetic algorithm + local search (GA+LS).

There are several processes shared among the three algorithms. For simplicity of notations, we assume the features of $\bx$ indexed by $D$ are the first $|D|$ features, i.e., $D=\{1,2,\dots,|D|\}$. Let $q \sim \mathcal{U}\{D\}$ represent a uniformly distributed random variable over $D$ indicating the indexical position of feature vector $\bx_D$ that will be perturbed. Perturbations to feature $\bx_q$ occur according to a standard normal distribution 
\begin{eqnarray}
\label{hnorm}
b_q \sim \psi(\sigma_q)= \frac{1}{ \sigma_q \sqrt{2 \pi}}  \textrm{exp} \left(-\frac{b_q^2}{2 \sigma_q^2} \right)
\end{eqnarray}
where $b_q$ is random variable representing the perturbation that occurs at indexical position $q$ and $\sigma_q$ is the standard deviation of feature $q$ obtained from the training data. Let $\mathbf{e}_q\in\mathbb{R}^{|D|}$ be a vector that equals one in the $q$th coordinate and zero in other places so that the perturbed version of $\bx_D$ is denoted by $\bx_D+b_q\mathbf{e}_q$. Let $[\bZ]_j$ represent the $j$th row of a matrix $\bZ$. Two additional shared parameters include $m$ which we will use to denote the total population size and $MaxIters$ which we will use to denote the number of iterations until an algorithm terminates.

\subsubsection{Hill-climbing + local search}

Our hill-climbing + local search (HC+LS) algorithm is based on that outlined by Mannino and Koushik \citep{Mannino2000} and is related by Algorithm \ref{algo:hillclimb} which calls a local search procedure, outlined in Algorithm \ref{algo:locsear}. In this algorithm, the best current solution, denoted by $\bx_D^*$, is perturbed a single feature $q$ at a time in order to find a better solution. There are $m$ single-feature perturbations that occur at each iteration, leading to $m$ perturbed versions of $\bx_D^*$, denoted by $\bx_D^*+b_{q_j}\mathbf{e}_{q_j}$, for $j=1,2,\dots,m$. We use Algorithm~\ref{algo:Proj} to convert the direction $\bx_D^*+b_{q_j}\mathbf{e}_{q_j}-\bar\bx_D$ into a feasible state and update $\bx_D^*$ along the direction that yields the smallest $g(\bx_D^*+b_{q_j}\mathbf{e}_{q_j}-\bar\bx_D)$, where $g$ is defined in \eqref{FeatureOptGenReform}.

\begin{algorithm}
	\caption{LS $\textbf{Local}(\bx_D)$}
	\label{algo:locsear}
	\begin{algorithmic}[1]
		\REQUIRE $\bx_D \in \mathbb{R}^{|D|}, (\bar{\bx}_U, \bar{\bx}_D) \in \mathbb{R}^{|U|+|D|}$, $\{c_i^+\}_{i\in D}$, $\{c_i^-\}_{i\in D}$, $\{l'_i\}_{i\in D}$, $\{u'_i\}_{i\in D}$, $B$, and $m$
        \FOR{$j=1$ to $m$}
        \STATE Generate $q_j \sim \mathcal{U}\{D\}$ and $b_{q_j}$ as \eqref{hnorm};
		\STATE $\bz_j = \textbf{Proj}_{\Delta_{D}}(\bx_D+b_{q_j}\mathbf{e}_{q_j} - \bar{\bx}_D)$;
        \ENDFOR
		\IF {$\textrm{min}\{g(\bz_j ); j=1,\dots,m\} < g(\bx_D-\bar\bx_D)$}
		\STATE $\bx_D= \bx_D+\bz_{j^{\prime}}\text{ with } j^{\prime}=\textrm{argmin}_{j=1,\dots,m}g(\bz_j)$
		\ENDIF
		\ENSURE $\bx_D$
	\end{algorithmic}
\end{algorithm}

We note here that the difference between regular HC and HC+LS is that HC operates on a first improvement basis, whereas HC+LS operates on a best improvement basis.

\begin{algorithm}
	\caption{HC+LS $\textbf{Hill}(\bar\bx)$}
	\label{algo:hillclimb}
	\begin{algorithmic}[1]
		\REQUIRE $(\bar{\bx}_U, \bar{\bx}_D) \in \mathbb{R}^{|U|+|D|}$, $\{c_i^+\}_{i\in D}$, $\{c_i^-\}_{i\in D}$, $\{l'_i\}_{i\in D}$, $\{u'_i\}_{i\in D}$, $B$, $m$, and $MaxIters$
		\STATE Initialize $\bx^{*}_D=\bar{\bx}_{D}$;
		\FOR{$iters=1$ to $MaxIters$}
		\STATE $\bx_D^{*}= \textrm{\textbf{Local}}(\bx_D^{*})$
		\ENDFOR
		\ENSURE $\bx^*_D$
	\end{algorithmic}
	
\end{algorithm}

\subsubsection{Genetic algorithm}

Genetic algorithms are composed of four primary processes: initial population generation, crossover, carryover, and mutation. Our real-valued genetic algorithm (GA) is outlined by Algorithm \ref{algo:genalg}. Prior to outlining such a method, we first relate the four aforementioned components.

At the first iteration of our genetic algorithm, an initial population is generated. For $j=1,2,\dots,m$, let $t_j \sim U\{D\}$ be a discrete uniform random variable. We then generate $q_{k,j} \sim \mathcal{U}\{D\}$ and $b_{q_{k,j}}$ as in \eqref{hnorm} for $k=1,2,\dots,t_j$ and define 
\begin{eqnarray}
\label{eq:upvecb}
\bb_j = \sum_{k=1}^{t_j}b_{q_{k,j}}\mathbf{e}_{q_{k,j}}; j=1,\dots,m.
\end{eqnarray}
We use $\bar{\bx}_D+\textbf{Proj}_{\Delta_{D}}(\bb_j)$ for $j=1,\dots,m$ as the initial population and store them as the rows of a $m\times |D|$ matrix $\bX_D^{\textrm{pop}}$, i.e,
\begin{align}
\label{eq:genpop}
[\bX_D^{\textrm{pop}}]_j = \bar{\bx}_D+\textbf{Proj}_{\Delta_{D}}(\bb_j); j=1,\dots,m.
\end{align}
We note that $\bb$ is updated $m$ times, resulting in unique entries in $\bX_D^{\textrm{pop}}$. Here, we apply \eqref{proj} to ensure that all population chromosomes are feasible.	

Following this, a simple procedure $\textrm{\textbf{Order}}(\bX_D^{\textrm{pop}})$ is called. This orders the rows $\bX_D^{\textrm{pop}}$ by objective function value from smallest to largest. Let $\beta \in (0,1]$ be a user specified parameter that denotes the proportion of the population that will be bred to produce the offspring for the next generation. We make a copy of the first $\left \lceil{m\beta}\right \rceil$ rows of $\bX_D^{\textrm{pop}}$ and store them as a $\left \lceil{m\beta}\right \rceil\times |D|$ matrix $\bX^{\textrm{cross}}_D$ with $[\bX^{\textrm{cross}}_D]_j = [\bX_{D}^{\textrm{pop}}]_j$ for $j=1,2,...,\left \lceil{m\beta}\right \rceil$. We then randomly shuffle the rows of $\bX^{\textrm{cross}}_D$ using a procedure $\textrm{\textbf{Shuffle}}(\bX_D^{\textrm{cross}})$.

Let $1-\gamma$ be the proportion of the population that should be composed of children ($\gamma$ being the proportion of the population that will be carried over, discussed shortly). We construct a vector of indices $\vartheta
\in\mathbb{N}^{\left \lceil{m\beta}\right \rceil}$ as
\begin{align}
\label{eq:crindset}
\vartheta = \left\{
\begin{array}{lc}
(1,2,\dots,\left \lceil{m(1-\gamma)}\right \rceil) & \textrm{if }(1-\gamma) \leq \beta \\
(1,2,\dots,\left \lceil{m\beta}\right \rceil,  & \textrm{otherwise}\\
r_{\left\lceil{m\beta}\right \rceil+1},r_{\left\lceil{m\beta}\right \rceil+2},\dots,r_{\left\lceil{m(1-\gamma)}\right \rceil})&
\end{array}\right.,
\end{align}
where $r_k \sim \mathcal{U} \{1,2,\dots,\left\lceil{m\beta}\right \rceil\}$ is a uniformly distributed random index for $k=\left\lceil{m\beta}\right \rceil+1,\left\lceil{m\beta}\right \rceil+2,\dots,\left\lceil{m(1-\gamma)}\right \rceil$.

Selected chromosomes are bred using single-point crossover outlined in Michalewicz, 2013 \citep{Michalewicz2013}, adapted to maintain feasibility via our projection operator. Without loss of generality, we assume $\left \lceil{m(1-\gamma)}\right \rceil$ is an even number and $\left \lceil{m(1-\gamma)}\right \rceil=2K$ for some integer $K$. For $k=1,2,\dots,K$, we use the vector $\vartheta$ defined in \eqref{eq:crindset} to create children from the matrix of parent chromosomes $\bX_D^{\textrm{cross}}$ by doing
\begin{align}
\small
\label{eq:crossover}
\begin{array}{l}
\bx_D^{2k-1} = \textrm{\textbf{Mut}}(([\bX_D^{\textrm{cross}}]_{\vartheta_{2k-1}}^1,...,[\bX_D^{\textrm{cross}}]_{\vartheta_{2k-1}}^{q_k-1},[\bX_D^{\textrm{cross}}]_{\vartheta_{2k}}^{q_k},\\\hspace{5cm} ...,[\bX_D^{\textrm{cross}}]_{\vartheta_{2k}}^{|D|}))\\
\bx_D^{2k} = \textrm{\textbf{Mut}}(([\bX_D^{\textrm{cross}}]_{\vartheta_{2k}}^1,...,[\bX_D^{\textrm{cross}}]_{\vartheta_{2k}}^{q_k-1},[\bX_D^{\textrm{cross}}]_{\vartheta_{2k-1}}^{q_k},\\ \hspace{5cm}...,[\bX_D^{\textrm{cross}}]_{\vartheta_{2k-1}}^{|D|}))
\end{array}
\end{align}
\normalsize
where $[\bZ]_j^i$ represent the entry in the $j$th row and the $i$th column of a matrix $\bZ$, 
$q_k \sim\mathcal{U}\{D\}$ is a selected crossover point, generated for a pair of parents, i.e., the rows $\vartheta_{2k-1}$ and $\vartheta_{2k}$ of $\bX_D^{\textrm{cross}}$ for $k=1,2,\dots,K$. \textbf{Mut}$(\bx_D)$ is the mutation operator defined as
\begin{eqnarray}
\label{eq:mutation}
\textbf{Mut}(\bx_D)=\bx_D+ I_{v} b_q\mathbf{e}_q,
\end{eqnarray}
where $q \sim \mathcal{U}\{D\}$, $b_{q}$ is generated as \eqref{hnorm}, $I_v$ is a binary random variable which equals zero and one with a probability of $v$ and $1-v$ respectively ($v$ is a user-specified parameter representing the probability of mutation occurring to allele $i$ by amount $b_i$ defined in \eqref{hnorm}).
Subsequently, the children are ensured feasible by
\begin{eqnarray}
\label{eq:feaschild}
\begin{array}{c}
\bx_D^{2k-1\prime} = \textrm{\textbf{Proj}}_{\Delta_{D}}(\bx_D^{2k-1} - \bar{\bx}_D) + \bar{\bx}_D\\
\bx_D^{2k\prime} = \textrm{\textbf{Proj}}_{\Delta_{D}}(\bx_D^{2k}- \bar{\bx}_D) + \bar{\bx}_D.
\end{array}
\end{eqnarray}
These feasible children are then stored as rows of a $\left \lceil{m(1-\gamma)}\right \rceil\times |D|$ matrix $\bX_D^{\textrm{child}}$.

The carryover procedure uses roulette wheel selection \citep{Michalewicz2013} to select chromosomes from the current generation that will survive to the next. Chromosomes that have larger (i.e., better) fitness values (where fitness denotes solution quality) have a higher likelihood of surviving to the next generation.

First, we create an inverted solution vector $\mathcal{P}\in\mathbb{R}^m$. These inverted solutions are transformed from  $g(\bx_D-\bar\bx_D)$ by function $\Lambda:\mathbb{R}^{|D|}\rightarrow\mathbb{R}$ defined as
\begin{eqnarray}
\label{eq:invsol}
\Lambda(\bx_D) = \omega - g(\bx_D-\bar\bx_D)
\end{eqnarray}

where $\omega$ is the worst-case solution possible and is assumed to be positive. Using $\Lambda(\cdot)$ we construct a vector of selection probability

\begin{align}
\label{eq:selecvec}
\mathscr{P} = \bigg(\frac{\Lambda([\bX_D^{\textrm{pop}}]_1)}{\sum_{j=1}^{m} \Lambda([\bX_D^{\textrm{pop}}]_j)}
,\dots,
\frac{\Lambda([\bX_D^{\textrm{pop}}]_m)}{\sum_{j=1}^{m} \Lambda([\bX_D^{\textrm{pop}}]_j)} \bigg).
\end{align}
Intuitively, higher quality solutions have larger probability to be selected, since we have already ordered $\bX_D^{\textrm{pop}}$ by $g(\bx_D-\bar\bx_D)$ so that $\mathscr{P}_j> \mathscr{P}_{j+1}$.

Using \eqref{eq:selecvec} we select chromosomes from the population matrix $\bX_{D}^{\textrm{pop}}$ to be carried over to the next generation by
\begin{align}
\label{eq:carryover}
\bx_{D}^{\textrm{carry},k} = [\bX_{D}^{\textrm{pop}}]_j \text{ with a probability of }\mathscr{P}_j
\end{align}
for $k=1,2,\dots,\left\lfloor m\gamma \right\rfloor$. Those selected children are stored as the $\left\lfloor m\gamma \right\rfloor\times |D|$ carryover matrix $\bX_D^{\textrm{carry}}$.

Using \eqref{eq:genpop},\eqref{eq:crossover}, \eqref{eq:mutation} and \eqref{eq:carryover} we construct our GA as outlined by Algorithm \ref{algo:genalg}. The procedure begins by initializing the best solution to the unperturbed chromosome $\bar{\bx}_D$. The algorithm then begins iteration, executing $MaxIters$ times. If it is the first iteration, the initial population is generated. The current population $\bX_D^{\textrm{pop}}$ is then evaluated and if a better solution is found, it is updated. Following this, a simple procedure $\textrm{\textbf{Order}}(\bX_D^{\textrm{pop}})$ is called. This orders $\bX_D^{\textrm{pop}}$ by objective function value from smallest to largest. Crossover points are then selected in an elitist fashion from this ordered matrix of chromosomes. Selected chromosomes are then randomly shuffled, using procedure $\textrm{\textbf{Shuffle}}(\bX_D^{\textrm{cross}})$, before crossover is applied to create the offspring chromosomes. Next, the carryover chromosomes are selected. Finally, the children and carryover chromosome matrices are concatenated to form the population for the next generation.

\begin{algorithm}
	\caption{GA $\textbf{Gen}(\bar{\bx})$}
	\label{algo:genalg}
	\begin{algorithmic}[1]
		\REQUIRE $(\bar{\bx}_U, \bar{\bx}_D) \in \mathbb{R}^{|U|+|D|}$, $\{c_i^+\}_{i\in D}$, $\{c_i^-\}_{i\in D}$, $\{l'_i\}_{i\in D}$, $\{u'_i\}_{i\in D}$, $B$, $m$, $MaxIters$, $\alpha$, $\beta$, $\gamma$, $v$ and $\omega$
		\STATE Initialize $\bx^{*}_D=\bar{\bx}_D$
		\FOR{$iters=1$ to $MaxIters$}
		\IF {$iters == 1$}
		\STATE Generate an initial population $\bX_D^{\textrm{pop}}$ using \eqref{eq:genpop}
		\ENDIF
\IF {$\textrm{min}\{g([\bX_D^{\textrm{pop}}]_j-\bar{\bx}_D ); j=1,\dots,m\} < g(\bx^{*}_D-\bar\bx_D)$}
		\STATE $\bx^*_D= [\bX_D^{\textrm{pop}}]_{j^{\prime}}\text{ with } j^{\prime}=\textrm{argmin}_{j=1,\dots,m}g([\bX_D^{\textrm{pop}}]_j-\bar{\bx}_D )$
		\ENDIF
		\STATE $\textrm{\textbf{Order}}(\bX_D^{\textrm{pop}})$
		\STATE $[\bX^{\textrm{cross}}_D]_j = [\bX_{D}^{\textrm{pop}}]_j$, $j=1,2,...,\left\lceil m\beta \right\rceil$
		\STATE$\textrm{\textbf{Shuffle}}(\bX_D^{\textrm{cross}})$
		\STATE Obtain $\bX_D^{\textrm{child}}$ from $\bX^{\textrm{cross}}_D$ according to \eqref{eq:crossover}
		\STATE Obtain $\bX_D^{\textrm{carry}}$ from $\bX_{D}^{\textrm{pop}}$ by \eqref{eq:carryover}
		\STATE Create the new population 
$
\bX_D^{\textrm{pop}} = \left(\begin{array}{c}\bX^{\textrm{child}}_D \\  \bX^{\textrm{carry}}_D\end{array}\right)
$
		\ENDFOR
		\ENSURE $\bx^{*}_D$
	\end{algorithmic}
\end{algorithm}

\subsubsection{Genetic algorithm + local search}

The third method is a genetic algorithm + local search (GA+LS). It is related by Algorithm \ref{algo:genalglocsear}. There are a few important distinctions between the original GA and that with local search applied. First, we reformulate the crossover procedure outlined by \eqref{eq:crossover} to be
\begin{align}
\label{eq:crossoverls}
\begin{array}{l}
\bx_D^{2k-1} = ([\bX_D^{\textrm{cross}}]_{\vartheta_{2k-1}}^1,...,[\bX_D^{\textrm{cross}}]_{\vartheta_{2k-1}}^{q_k-1},[\bX_D^{\textrm{cross}}]_{\vartheta_{2k}}^{q_k},\\ \hspace{5cm}...,[\bX_D^{\textrm{cross}}]_{\vartheta_{2k}}^{|D|})\\
\bx_D^{2k} = ([\bX_D^{\textrm{cross}}]_{\vartheta_{2k}}^1,...,[\bX_D^{\textrm{cross}}]_{\vartheta_{2k}}^{q_k-1},[\bX_D^{\textrm{cross}}]_{\vartheta_{2k-1}}^{q_k},\\ \hspace{5cm}...,[\bX_D^{\textrm{cross}}]_{\vartheta_{2k-1}}^{|D|})
\end{array}
\end{align}
for $k=1,2,\dots,K$.
The reader will note that here the mutation procedure is not applied.

Second, we incorporate the use of the local search (LS) procedure previously outlined in Algorithm \ref{algo:locsear}. Here, we set parameter $m$ equal to $\xi$, which dictates the extent of the search.


GA+LS is outlined by Algorithm \ref{algo:genalglocsear}. The differences between this method and the original GA are outlined in \textcolor{blue}{blue}. At line 14 the LS procedure is applied to each of the non-mutated children. The best solution obtained from LS is the child chromosome that is kept for the next generation.

\begin{algorithm}[h]
	\caption{GA+LS $\textbf{GenLoc}(\bar{\bx})$}
	\label{algo:genalglocsear}
	\begin{algorithmic}[1]
		\REQUIRE $(\bar{\bx}_U, \bar{\bx}_D) \in \mathbb{R}^{|U|+|D|}$, $\{c_i^+\}_{i\in D}$, $\{c_i^-\}_{i\in D}$, $\{l'_i\}_{i\in D}$, $\{u'_i\}_{i\in D}$, $B$, $m$, $MaxIters$, $\alpha$, $\beta$, $\gamma$, $v$ and $\xi$
		\STATE Initialize $\bx^{*}_D=\bar{\bx}_D$
		\FOR{$iters=1$ to $MaxIters$}
		\IF {$iters == 1$}
		\STATE Generate an initial population $\bX_D^{\textrm{pop}}$ using \eqref{eq:genpop}
		\ENDIF
\IF {$\textrm{min}\{g([\bX_D^{\textrm{pop}}]_j-\bar{\bx}_D ); j=1,\dots,m\} < g(\bx^{*}_D-\bar\bx_D)$}
		\STATE $\bx^*_D= [\bX_D^{\textrm{pop}}]_{j^{\prime}}\text{ with } j^{\prime}=\textrm{argmin}_{j=1,\dots,m}g([\bX_D^{\textrm{pop}}]_j-\bar{\bx}_D )$
		\ENDIF
		\STATE $\textrm{\textbf{Order}}(\bX_D^{\textrm{pop}})$
		\STATE $[\bX^{\textrm{cross}}_D]_j = [\bX_{D}^{\textrm{pop}}]_j$, $j=1,2,...,\left\lceil m\beta \right\rceil$
		\STATE $\textrm{\textbf{Shuffle}}(\bX_D^{\textrm{cross}})$
		\STATE \textcolor{blue}{Obtain} $\mathcolor{blue}{\bX^{\textrm{child}}_D}$ \textcolor{blue}{from} $\mathcolor{blue}{\bX^{\textrm{cross}}_D}$ \textcolor{blue}{by \eqref{eq:crossoverls}}
		\STATE $\mathcolor{blue}{[\bX^{\textrm{child}\prime}_D]_j = \textrm{\textbf{Local}}([\bX^{\textrm{child}}_D]_j),j=1,...,\left\lceil m(1-\gamma) \right\rceil}$
		\STATE Obtain $\bX^{\textrm{carry}}_D$ from $\bX_D^{\textrm{pop}}$ by \eqref{eq:carryover}
		\STATE Create the new population
$
\mathcolor{blue}{
\bX_D^{\textrm{pop}} = \left(\begin{array}{c}\bX^{\textrm{child}\prime}_D \\  \bX^{\textrm{carry}}_D\end{array}\right)
}
$
		\ENDFOR
		\ENSURE $\bx_D^*$
	\end{algorithmic}
\end{algorithm}

\subsection{Sensitivity analysis-based methods}

As discussed in Section 2, sensitivity analysis is closely related to inverse classification. Therefore, we propose two sensitivity analysis-based algorithms that serve as baselines against which the heuristic-based methods can be compared against. \textcolor{black}{To our knowledge, no past methods addressing this problem have been proposed. Therefore, we craft these ourselves, and believe that they represent a reasonable initial attempt at a solution.} Such methods can be viewed as a combination of local and variable perturbation methods of sensitivity analysis.

We refer to the first sensitivity analysis-based method as Local Variable Perturbation--Best Improvement (LVP-BI). This method calls for perturbing a single feature $i \in D$ to the extent of feasibility given by $\textrm{min}\big\{\frac{B}{c_i}, u^{\prime}_i \big\}$. The single feature perturbation having the greatest objective function improvement is the one that is accepted. If some budget remains following this perturbation, subsequent perturbations are performed (e.g.,~double feature, triple feature, etc. perturbations). 




Our second method, which we refer to as Local Variable Perturbation--First Improvement (LVP-FI), is very similar to that of LVP--BI. Instead of accepting the best perturbation over all $i\in D$ it accepts the first perturbation that leads to a better objective function value, where $i$ is selected at random.



\section{Experiments}

In this section we first outline our choices regarding the parameters of the inverse classification framework and then apply our methods to two freely available datasets. Our experiments will evaluate the five methods by examining the average likelihood of test instances conforming to a non-ideal class over varying budget constraints.. First, we will explore the capability of each algorithm in reducing the likelihood of test instances conforming to a non-ideal class. Additionally, we will examine the perturbations made to an individual test instance, selected at random, by the top performing algorithm. \textcolor{black}{We wish to emphasize that practical and real-world use of these methods should be undertaken with experts in the domain of use. We further emphasize that inverse classification puts the individual at the center of the process and optimizes over his/her current values. Therefore, if an individual so choses, he/she can adjust expert-specified costs according to their own outlook on what may be more or less difficult to change.}

\subsection{Experiment Parameters and Evaluation}

There are three choices that need to be made regarding the established inverse classification framework: the learning algorithm, the indirectly changeable feature estimator, and the method we will use to set the lower- and upper-bounds that directly changeable features can take.

\subsubsection{Objective Function}
We selected the Random forest classifier \citep{Breiman2001} to evaluate each of the five methods. We chose this as it is (a) an ensemble classifier and (b) composed of weak-learner decision trees. Both (a) and (b) are separately non-differentiable, and comprehensively help highlight the need for the GIC formulation we have proposed. The returned objective function value will be the proportion of decision trees in the ensemble voting in favor of the class to be minimized. As such $f(\cdot) \in [0,1]$. We therefore can also parameterize $\omega=1$, the worst-case objective function value in \eqref{eq:invsol}.

\subsubsection{Indirectly Changeable Feature Estimation}

The inverse classification framework allows for any smooth model $H(\cdot)$ to be selected to estimate the indirectly changeable features. We elect to use a kernel regression method \citep{Nadaraya1964,Watson1964}
\begin{eqnarray}
\label{kernelreg}
\bx_I &=& \frac{\sum_{i=1}^{n} k([\bx^i_D, \bx^i_U],[\bx_D, \bx_U])\bx_I^i}{\sum_{i=1}^{n} k([\bx^i_D, \bx^i_U],[\bx_D, \bx_U])},
\end{eqnarray}

\noindent where $\bx^i$ is a training instance and $k(\bx,\bx')=\exp\left(-\frac{\|\bx-\bx'\|^2}{2\sigma^2}\right)$ is the Gaussian kernel. We elect to use this function and corresponding Gaussian kernel for its similarity-based estimation properties. We cross-validate this model on each of the indirectly changeable features in order to learn the best $\sigma$ for each.

\subsubsection{Bound-setting method and cost function}

Lash et al.,~\citep{Lash2016} outline two methods of specifying lower- and upper-bounds for the directly changeable features. Each result in different algorithmic behavior. In our experiments we use the Hard-line bound-setting method. Under this method we specify, for feature $i \in D$, the upper- and lower- bounds such that $i$ can only either increase or decrease. If feature $i$ should increase from its current value of $\bx_i$ we set $l_i = \bar{\bx}_i$. If feature $i$ should decrease from it's current value of $\bx_i$ we set $u_i = \bar{\bx}_i$. This allows us to maintain more control over what we know and believe to be the beneficial direction of feature movement. We do note, however, that under different circumstances (e.g., uncertainty) it may be beneficial to allow the optimization to learn the most beneficial direction of feature movement.

In this set of experiments we elect to explore the effects of non-linear costs, related by \ref{FeasibleSetQuadratic}. We elect to do so as non-linear costs, to the best of our knowledge, have not been explored in past works.

\subsubsection{Evaluating Recommendations}

To evaluate the success of the inverse classification we use an established procedure originally outlined in \citep{Chi2012} and refined in \citep{Lash2016}. This process entails initially splitting a dataset $\mathbb{D}$ randomly into two equal parts $\mathbb{D}^{\textrm{train}}$, where the first is used for training the random forest model upon which inverse classification will take place. The second set $\mathbb{D}^{\textrm{ho}}$, is the held-out set of data to which inverse classification will be applied.

$\mathbb{D}^{\textrm{ho}}$ is further partitioned into $k$ distinct subsets which we can denote $\mathbb{D}^{\textrm{ho}}_i$, $i=1,...,k$ ($k=10$ in our experiments). The process of evaluation entails that we perform inverse classification on $\bar{\bx} \in \mathbb{D}^{\textrm{ho}}_i$ and use $\{\mathbb{D}^{\textrm{ho}}\} \setminus \{\mathbb{D}^{\textrm{ho}}_i\}$ to train a separate model to evaluate the success of the inverse classification. Such a process ensures that no information used to perform the inverse classification and obtain recommendations is used in evaluating how successful the process actually was. \textcolor{black}{Additionally, this helps ensure that the classifier used to make the recommendations has not overfit the data.}

\subsection{Student Performance: Grade-improving recommendations}

Our first set of experiments are conducted on a UCI Machine Learning Repository dataset called Student Performance \citep{cortez2008}. This dataset consists of Portuguese students enrolled in two different classes: a math class and a Portuguese language class. Represented as two disjoint, but overlapping datasets, we elect to use the Portuguese language set as it has the larger number of instances ($n=649$).

\subsubsection{Data Description}

Each individual in Student Performance is initially represented by 45 features, including a unique identifier (discarded) and class variable $y$, which we define to be whether or not a student's final grade was above a C ($y=0$) or, conversely, less than or equal to a C ($y=1$). Our GIC methods will attempt to reduce the likelihood of earning a grade of C or worse. We discard the two intermediary grade reports to reflect a long-term goal of earning a higher grade overall and make the problem more realistic. The full set of features and corresponding parameters can be viewed in the Supplemental Material.



The parameters set for the three heuristic-based methods in these experiments are related by Table \ref{tab:benchparam}, as is the computational complexity. We arrived at these after a brief exploration of the parameter space, selecting values that were comparable so that performance could be equivalently compared. For GA+LS, we kept $MaxIters$ (abbreviated $MI$) lower because of the added complexity of the $\xi$ parameter.

\begin{table}
	\centering
	\begin{tabular}{|l|c|c|c|}
		\hline
		Param & HC + LS & GA & GA + LS \\
		\hline
		$MI$ & 300 & 300 & 150 \\
		\hline
		$m$ & 15 & 15 & 15 \\
		\hline
		$\alpha$ & NA & $.30$ & NA \\
		\hline
		$\beta$ & NA & $.40$ & $.40$ \\
		\hline
		$\gamma$ & NA & $.10$ & $.10$ \\
		\hline
		$\xi$ & NA & NA & $6$ \\
		\hline
		Big $O$ & $MI*m$ & $MI*m$  & $MI*m*\xi$ \\
		\hline
	\end{tabular}
	\caption{Benchmark experiment parameters \label{tab:benchparam}}
\end{table}

\subsubsection{Results}

\begin{figure}[t]
	\centering
	\includegraphics[width=.62\linewidth]{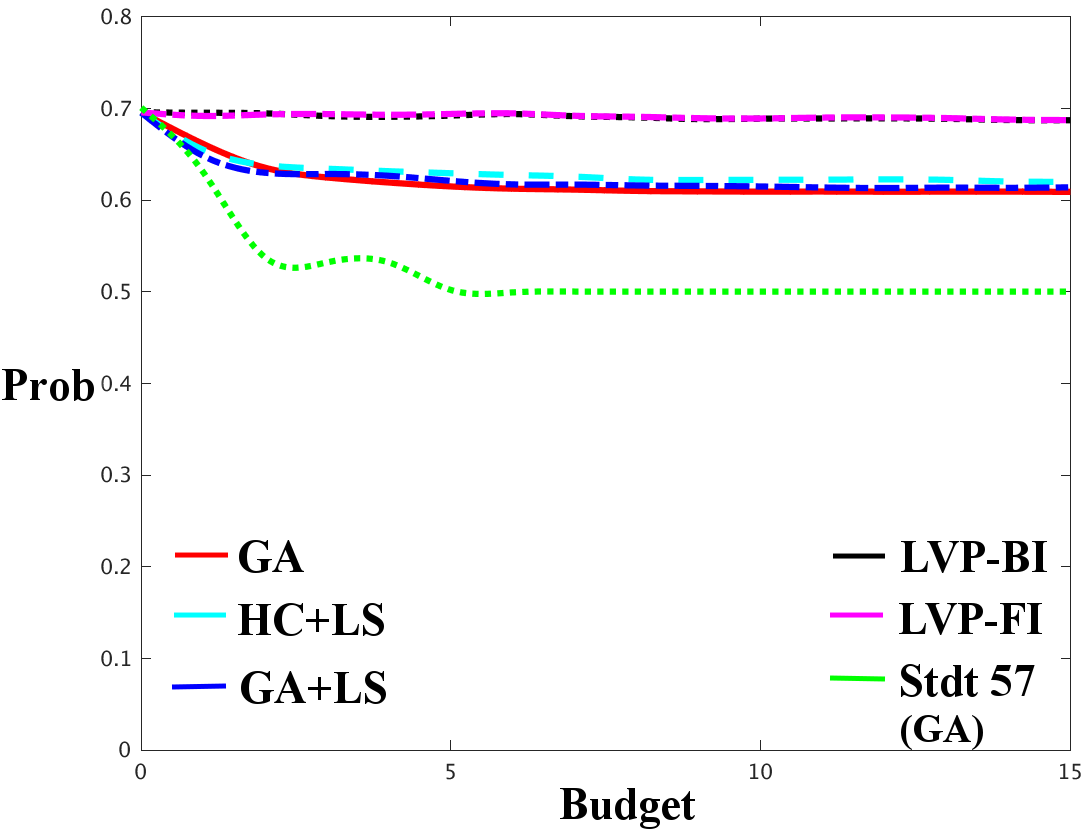}
	\caption{The average predicted probability vs budget for three heuristic- and two sensitivity analysis-based methods. GA result for Student 57.}
	\label{fig:bench_all_prop}

\end{figure}

We first examine the success of reducing the average predicted probability for each of the five methods.  These results are reported in Figure \ref{fig:bench_all_prop}. We report each over 15 increasing budgetary constraints. Additionally, we include the best result on a randomly selected positively classified instance -- Student 57 -- obtained using GA.

As we can observe in Figure \ref{fig:bench_all_prop} the two sensitivity analysis-based methods were unsuccessful. The result also shows that the three heuristic-based methods are comparable, with GA and GA+LS declining slightly faster than HC+LS. We include more detailed information about the performance of each method in the Supplemental Materials. 



\begin{figure}[t]
	\centering
	\includegraphics[width=.62\linewidth]{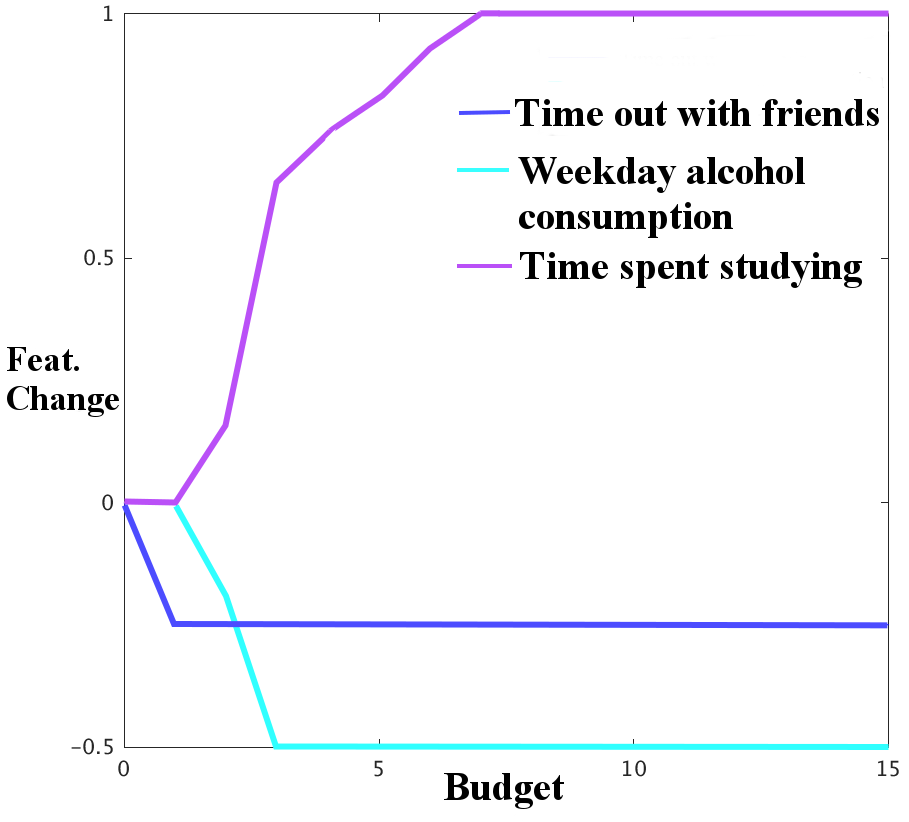}
	\caption{GA recommended changes to Student 57.}
	\label{fig:recgabench_all}
	
\end{figure}


We report the changes made to ``Student 57'' in Figure \ref{fig:recgabench_all} for the method most successful in reducing their predicted probability: GA. We report this so that the reader may have a better idea of what  such recommendations look like. GA recommends the student to increase study time and curb weekday alcohol consumption, as well as to decrease time out with friends.

\textcolor{black}{Cumulatively, the three heuristic methods were, on average, able to reduce the probability from approximately 70\% to 62\% at a budget level of three.  Individually, the best performing method was able to reduce Studet 57's probability from 70\% to 50\% at a budget level of five.}

\subsection{Cardiovascular disease mitigating lifestyle recommendations}

Our second set of experiments is conducted on a real-world patient dataset, derived from the ARIC study. These data are freely available upon request from BioLINCC.


\subsubsection{Data Description}

These data represent patients, for whom we have known cardiovascular disease (CVD) outcomes over a 10 year period. There are 110 defined features for each patient. Patients who, during the course of the 10 year period have probable myocardial infarction (MI), definite MI, suspect MI, definite fatal coronary heart disease (CHD), possible fatal CHD, or stroke have $y=1$ and $y=0$ otherwise. Patients who had a pre-existing CVD event are excluded from our dataset, giving us a total of $n=12007$ patients. This set of experiments is meant to more closely reflect a real-world scenario and, as such, is guided by a CVD specialist. The full list of features, their feature designation (e.g., changeable) and parameters (e.g.,~cost) can be viewed in the Supplemental Materials.

After a brief exploration of the parameter space, we arrived at the same set of parameters as in the previous experiment (Student Performance). We omit the duplicate table and refer to Table \ref{tab:benchparam}. Additionally, because of the size of the testing dataset, and the computational complexity associated with the heuristic-based methods, we elected to test on a subset of data. We used all 587 positive test instances and another 587 randomly selected negative test instances, giving us a final evaluative test set size of 1164. Evaluation models were constructed using the full set of data by the procedure outlined in Section 4.1.4.

\subsubsection{Results}

\begin{figure}
	\centering
	\includegraphics[width=.62\linewidth]{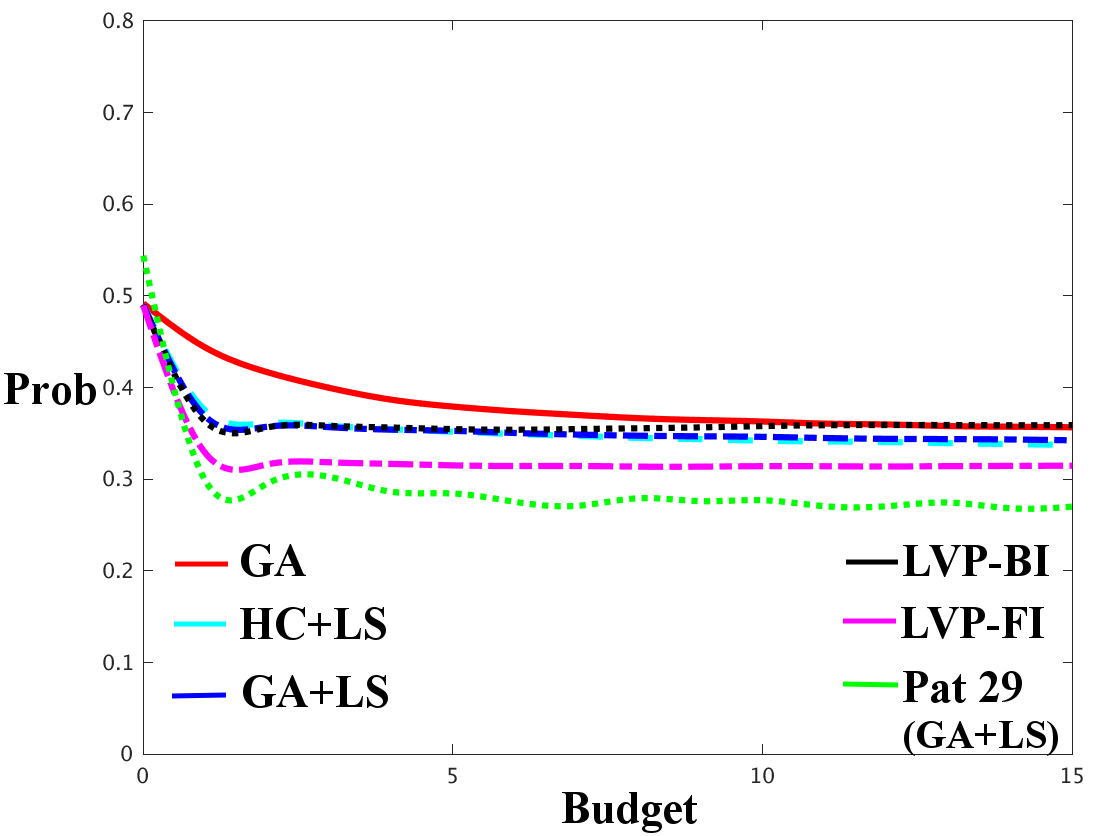}
	\caption{The average predicted probability vs budget for the three heuristic- and two sensitivity analysis-based methods. GA+LS result for Patient 29.}
	\label{fig:cvd_prop_all}
\end{figure}

We first examine the success of reducing the average predicted probability using the five outlined methods.  These results are reported in Figure \ref{fig:cvd_prop_all}. We report each over 15 increasing budgetary constraints. Additionally, we include the best result on a randomly selected positively classified instance -- Patient 29 -- obtained using GA+LS.

The results obtained for the heuristic-based methods are similar to those of Student Performance. There is a striking difference, however, between those and the sensitivity based-method results here. We observe that LVP-FI outperforms all other methods, while LVP-BI is comparable to GA and GA+LS. HC+LS performs the worst. The stark difference in performance of LVP-FI and LVP-BI on this dataset vs.~that of student performance may suggest that there are instances in which it is advantageous to use sensitivity analysis-based methods over those that are heuristic-based, and vice-versa. We leave such an analysis for future work.

\begin{figure}
	\centering
	\includegraphics[width=.65\linewidth]{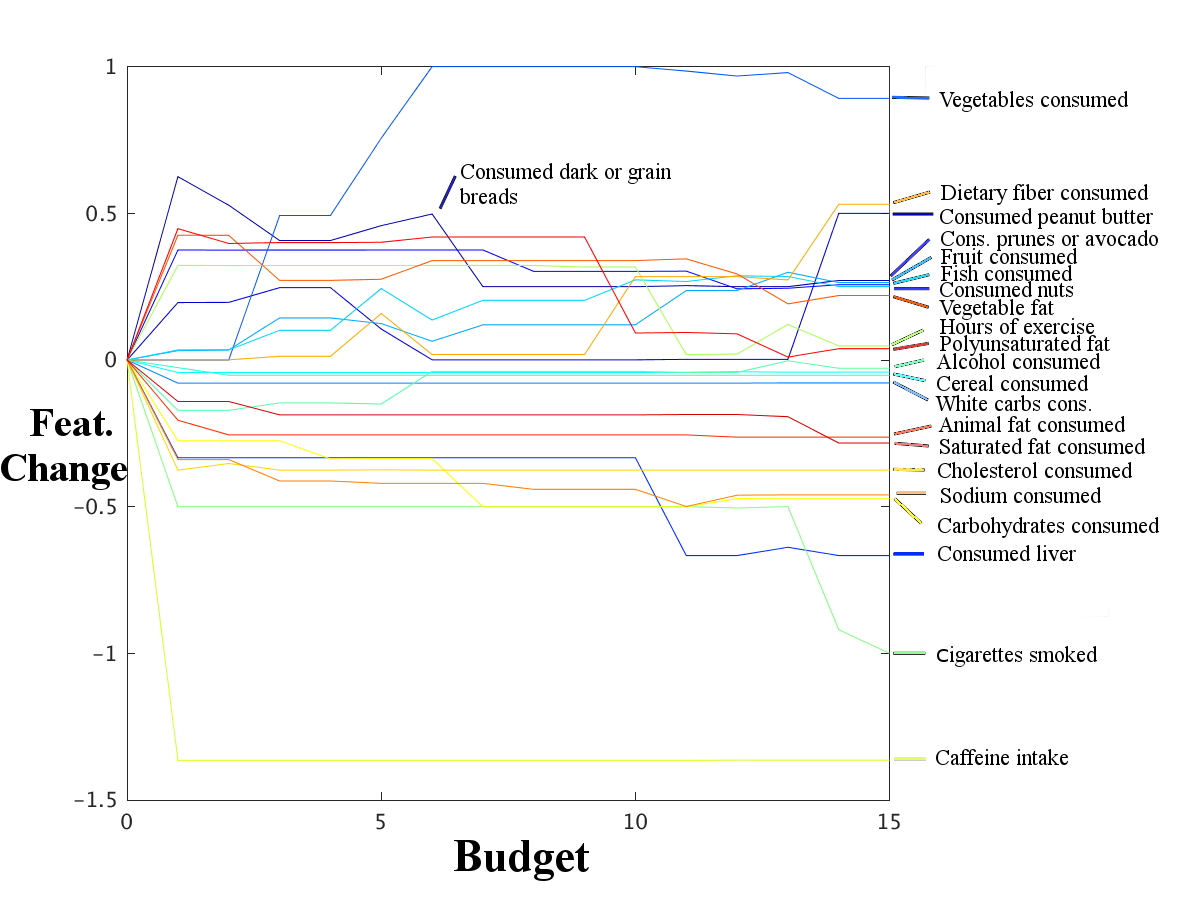}
	\caption{GA+LS recommended changes to Patient 29.}
	\label{fig:rec_all_cvd}
\end{figure}

We report the changes made to ``Patient 29'' in Figure \ref{fig:rec_all_cvd} for the method most successful in reducing the patient's predicted probability: GA+LS. Here we observe that the number of feature changes recommended are quite numerous: there are 22 of them. This suggests that it may be beneficial to include sparsity constraints.

\textcolor{black}{Cumulatively, these results show that, on average, risk can be taken from approximately 50\% to 30-35\%, depending upon the method, at a budgetary level of two. At the individual level, using the best method, Patient 29's risk can be lowered from 55\% to less than 30\%, also a at a budgetary level of two.}



\section{Conclusions}

In this work we propose and solve \textit{generalized inverse classification} by working backward through the previously un-navigable random forest classifier using five proposed algorithms that we incorporated into a framework, updated to account for non-linear costs, that leads to realistic recommendations. Future work is needed to analyze instances in which one method may outperform another, the performance of other classifiers and constraints limiting the number of features that are changed.

\setlength{\bibsep}{0pt plus 0.3ex}
\bibliography{lash_icdm_2016}

\begin{thebibliography}{10}

\bibitem{isukapalli1999}
S.~S. Isukapalli, {\em Uncertainty Analysis of Transport-transformation
  Models}.
\newblock PhD thesis, Citeseer, 1999.

\bibitem{Yao2003}
J.~Yao, ``Sensitivity analysis for data mining,'' in {\em Fuzzy Information
  Processing Society, 2003. NAFIPS 2003. 22nd International Conference of the
  North American}, pp.~272--277, July 2003.

\bibitem{Aggarwal2010}
C.~C. Aggarwal, C.~Chen, and J.~Han, ``{The inverse classification problem},''
  {\em Journal of Computer Science and Technology}, vol.~25, no.~May,
  pp.~458--468, 2010.

\bibitem{Chi2012}
C.~L. Chi, W.~N. Street, J.~G. Robinson, and M.~A. Crawford, ``{Individualized
  patient-centered lifestyle recommendations: An expert system for
  communicating patient specific cardiovascular risk information and
  prioritizing lifestyle options},'' {\em Journal of Biomedical Informatics},
  vol.~45, no.~6, pp.~1164--1174, 2012.

\bibitem{Yang2012}
C.~Yang, W.~N. Street, and J.~G. Robinson, ``{10-year CVD risk prediction and
  minimization via inverse classification},'' in {\em Proceedings of the 2nd
  ACM SIGHIT symposium on International health informatics - IHI '12},
  pp.~603--610, 2012.

\bibitem{Mannino2000}
M.~V. Mannino and M.~V. Koushik, ``{The cost minimizing inverse classification
  problem : A algorithm approach},'' {\em Decision Support Systems}, vol.~29,
  no.~3, pp.~283--300, 2000.

\bibitem{Barbella2009}
D.~Barbella, S.~Benzaid, J.~Christensen, B.~Jackson, X.~V. Qin, and
  D.~Musicant, ``{Understanding support vector machine classifications via a
  recommender system-like approach},'' in {\em Proceedings of the International
  Conference on Data Mining}, pp.~305--11, 2009.

\bibitem{Pendharkar2002}
P.~C. Pendharkar, ``A potential use of data envelopment analysis for the
  inverse classification problem,'' {\em Omega}, vol.~30, no.~3, pp.~243--248,
  2002.

\bibitem{Lash2016}
M.~T. Lash, Q.~Lin, W.~N. Street, and J.~G. Robinson, ``A budget-constrained
  inverse classification framework for smooth classifiers,'' {\em arXiv
  preprint; arxiv:1605.09068}, 2016.

\bibitem{Lash2016early}
M.~T. Lash and K.~Zhao, ``Early predictions of movie success: The who, what,
  and when of profitability,'' {\em Journal of Management Information Systems},
  vol.~33, no.~3, pp.~874--903, 2016.

\bibitem{Michalewicz2013}
Z.~Michalewicz, {\em Genetic algorithms+ data structures= evolution programs}.
\newblock Springer Science \& Business Media, 2013.

\bibitem{Breiman1996}
L.~Breiman, ``{Bagging Predictors},'' {\em Machine Learning}, vol.~24,
  pp.~123--140, 1996.

\bibitem{Freund1996}
Y.~Freund and R.~E. Schapire, ``{Experiments with a new boosting algorithm},''
  {\em Thirteenth International Conference on Machine Learning}, pp.~148--156,
  1996.

\bibitem{Breiman2001}
L.~Breiman, ``{Random forests},'' {\em Machine learning}, vol.~45, no.~1,
  pp.~5--32, 2001.

\bibitem{Nadaraya1964}
E.~a. Nadaraya, ``{On estimating regression},'' {\em Theory of Probability \&
  Its Applications}, vol.~9, no.~1, pp.~141--142, 1964.

\bibitem{Watson1964}
G.~S. Watson, ``{Smooth regression analysis},'' {\em The Indian Journal of
  Statistics, Series A}, vol.~26, no.~4, pp.~359--372, 1964.

\bibitem{cortez2008}
P.~Cortez and A.~M.~G. Silva, ``{Using data mining to predict secondary school
  student performance},'' in {\em Proceedings of 5th Annual Future Business
  Technology Conference}, EUROSIS, 2008.

\end{thebibliography}
\bibliographystyle{ieeetr}
\twocolumn[\section*{Supplemental Material -- Generalized Inverse Classification}]
\renewcommand{\figurename}{SF.}
\setcounter{figure}{0}
\renewcommand{\tablename}{ST.}
\setcounter{table}{0}

\subsection*{Supplementary Tables}
These tables show the unchangeable, indirectly changeable, and directly changeable features for each of our two freely available datasets. For each of the indirectly changeable features, the kernel regression $\sigma$ parameter is also included.

\begin{table}[!htbp]
	\centering
	\begin{tabular}{|C{6.4cm}|}
		\hline
		\textbf{Feature Name} \tabularnewline
		\hline
		School Attended, Sex, Age, Address, Size of family, Parent's cohabitation status, Mother's education, Father's education, Mother's job= "At Home", Mother's job="Health", Mother's job="Other", Mother's job="Services", Mother's job="Teacher", Father's job="Teacher", Father's job="Other", Father's job="Services", Father's job="Health", Father's job="At Home", Reason for school="Course", Reason for school="Other", Reason for school="home", Reason for school="Reputation", Guardian="Mother", Guardian="Father", Guardian="Other", Time spent traveling to school \tabularnewline
		\hline
	\end{tabular}
	\caption{Unchangeable features for the Student Performance dataset.}
	\label{tab:bench_unchange}
\end{table}

\begin{table}[!htbp]
	\centering
	\begin{tabular}{|C{6.4cm}|}
		\hline
		\textbf{Feature Name: $\sigma$} \tabularnewline
		\hline
		Extra-curricular activities: 1.5, Higher education aspirations: 1.0,
		In a romantic relationship: 1.5, Free time after school: 1.0 \tabularnewline
		\hline
	\end{tabular}
	\caption{Indirectly changeable features and learned kernel regression $\sigma$ parameters for the Student Performance dataset.}
	\label{tab:bench_indirchange}
\end{table}

\begin{table}[!htbp]
	\centering
	\begin{tabular}{|C{.8cm}|C{6.4cm}|}
		\hline
		$\mathbf{c^{+}/c^{-}}$ & \textbf{Feature:Cost} \tabularnewline
		\hline
		$c^{+}$ & Study time: 7, Paid tutoring: 8 \tabularnewline
		\hline
		$c^{-}$ & Time out with friends: 6, Weekday alcohol: 3, Weekend alcohol: 6, Absences from class: 5 \tabularnewline
		\hline
	\end{tabular}\protect\caption{Directly changeable variables for the Student Performance dataset.}
	\label{tab:bench_changeable}
\noindent\makebox[\linewidth]{\rule{\linewidth}{0.4pt}}
\end{table}

\begin{table}[!htbp]
	\centering
	\begin{tabular}{|C{6.4cm}|}
		\hline
		\textbf{Feature Name} \tabularnewline
		\hline
		Insulin (uu-ml), Height (cm), Age, Peripheral Artery Disease, Peripheral Artery Disease (definition 2), 
		Plaque/shadowing in either internal, Plaque in either internal carotid, Cholesterol lowering med (last 2 weeks),
		Hypertension (definition 5), Education level, Diabetes, Age when menopause began, Menopause status, Ever smoked cigarettes,
		High blood pressure med (past 2 weeks), Agina-chest pain med (past 2 weeks), Heart rhythm control med (past 2 weeks),
		Heart failure med (past 2 weeks), Blood thinning med (past 2 weeks), Blood sugar med (past 2 weeks), Stroke med (past 2 weeks),
		Walking leg pain med (past 2 weeks), Headache or cold med (past 2 weeks), Pain meds (past 2 weeks), Gender, Race, Years smoked cigarettes \tabularnewline
		\hline
	\end{tabular}
	\caption{Unchangeable features for the ARIC CVD dataset.}
	\label{tab:cvd_unchange}
\end{table}

\begin{table}[!htbp]
\centering
	\begin{tabular}{|C{6.4cm}|} 
		\hline
		\textbf{Feature Name: $\sigma$} \tabularnewline
		\hline
		BMI (Body Mass Index): .5, Recalibrated HDL cholesterol (mg/dl): .5, Re-calibrated LDL cholesterol (mg/dl): .5,Total cholesterol (mmol/L): .5, Total triglycerides (mmol/L): .5, 2nd and 3rd systolic blood pressure (avg.): .5, 2nd and 3rd systolic blood pressure (avg.) Num 2: .5, Waist girth (cm): .5, Hip girth (cm): .5, Heart rate: .5, White blood count: .5,
		Apolipoprotein AI(mg-dl): .5, Apolipoprotein B (mg-dl): .5, Apolp(A) Data (ug-ml): .5, Ankle-brachial index (Def 4): .5, FV(1)/FVC Predicted (\%): .25, FEV(1) (L): .5, FVC (L): .5, 
		Hematocrit: .5, Hemaglobin: .5, Platelet count: .5, Neutrophils: .5, Neutrophil bands: .5, Lymphocytes: .5, Monocytes: .5, Eosinophils: .5, Basophils: .5, APTT Value: .5, VIII: C Value: .5, Fibrinogen Value: .5, VII Value: .5, ATIII Value: .5, Protein: C Value: .5, VWF Value: .5 \tabularnewline 
		\hline
	\end{tabular}

\end{table}

\begin{table}[H]
	\centering
	\begin{tabular}{|C{6.4cm}|}
		\hline
		\textbf{Feature Name: $\sigma$} \tabularnewline
		\hline
		Cornell voltage (uV): .5, Waist-hip ratio: .5, Vegetable fat (\% kcal): .5, Carbs (\% kcal): .5, Alcohol (\% kcal): .5, Omega fatty acid (g): .5, Calf girth (cm): .5, Subcaps measure 2 (mm): .5, Triceps measure 2 (mm): .5, Uric acid (mg-dl): .5, Total protein (gm-dl): .5, Albium (gm-dl): .5, Phosphorus (mg-dl): .5, Magnesium (meq-l): .5, Calcium (mg-dl): .5, Urea nitgrogen (mg-dl): .5, Potassium (mmol-l): .5, 
		Sodium (mmol-l): .5, Creatinine (mg-dl): .5, Weight (lb): .5,Total fat (\% kcal): .5, Saturate fatty acid (\% kcal): .5, Protein (\% kcal): .5, Polyunsaturated fatty acid (\% kcal): .5, Monounsaturated fatty acid (\% kcal): .5, Total fat (g): .25 \tabularnewline
		\hline
	\end{tabular}
	\caption{Indirectly changeable features and learned kernel regression $\sigma$ parameters for the ARIC CVD dataset.}
	\label{tab:bench_indirchange}
\end{table}

\begin{table}[H]
	\centering
	\begin{tabular}{|C{.8cm}|C{6.4cm}|}
		\hline
		$\mathbf{c^{+}/c^{-}}$ & \textbf{Feature:Cost} \tabularnewline
		\hline
		$c^{+}$ & Dark or grain breads: 3, Peanut butter: 4, Nuts: 5, Other(prunes,avocado): 5, Vegetables: 6, Fruit: 6, Fiber: 7, Vegetable fat: 5, Polyunsaturated fat: 5 \tabularnewline
		\hline
		$c^{-}$ & Liver: 8, White carbs: 6, Fish: 9, Cereal: 4, Cigarettes: 9, Caffeine: 7, Carbs: 7, Cholesterol: 6, Sodium: 7, Animal fat: 7, Saturated fat: 6 \tabularnewline
		\hline
		$c^{+}/c^{-}$ & Exercise hours: 10, Alcohol: 9 \tabularnewline
		\hline
	\end{tabular}\caption{Directly changeable variables for the ARIC CVD dataset.}
	\label{tab:aric_changeable}
\noindent\makebox[\linewidth]{\rule{\linewidth}{0.4pt}}
\end{table}


\subsection*{Supplementary Figures}
These figures show additional algorithm-specific results that supplement and support certain conclusions that are made in the main content of the paper. Here, \textcolor{red}{red} shows the average probability, \textcolor{yellow}{yellow} shows the probability for a randomly selected instance and \textcolor{blue}{blue} shows the bottom 5 and top 95 \% of probabilities.

\begin{figure}[H]
	\centering
		\begin{subfigure}[!t]{\linewidth}
			\centering
			\includegraphics[width=\linewidth]{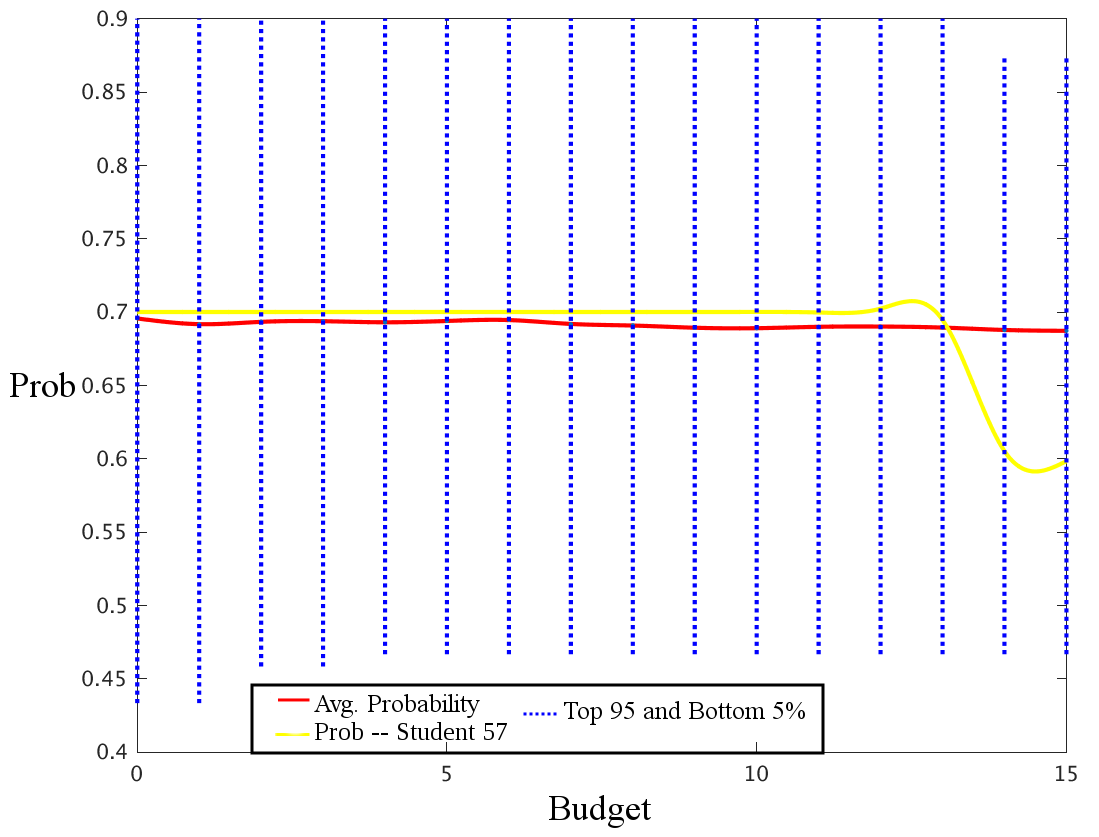}
			\caption{LVP-FI
				\label{fig:benchlvpfi_prob}}
		\end{subfigure}\par
		\begin{subfigure}[t]{\linewidth}
			\centering
			\includegraphics[width=\linewidth]{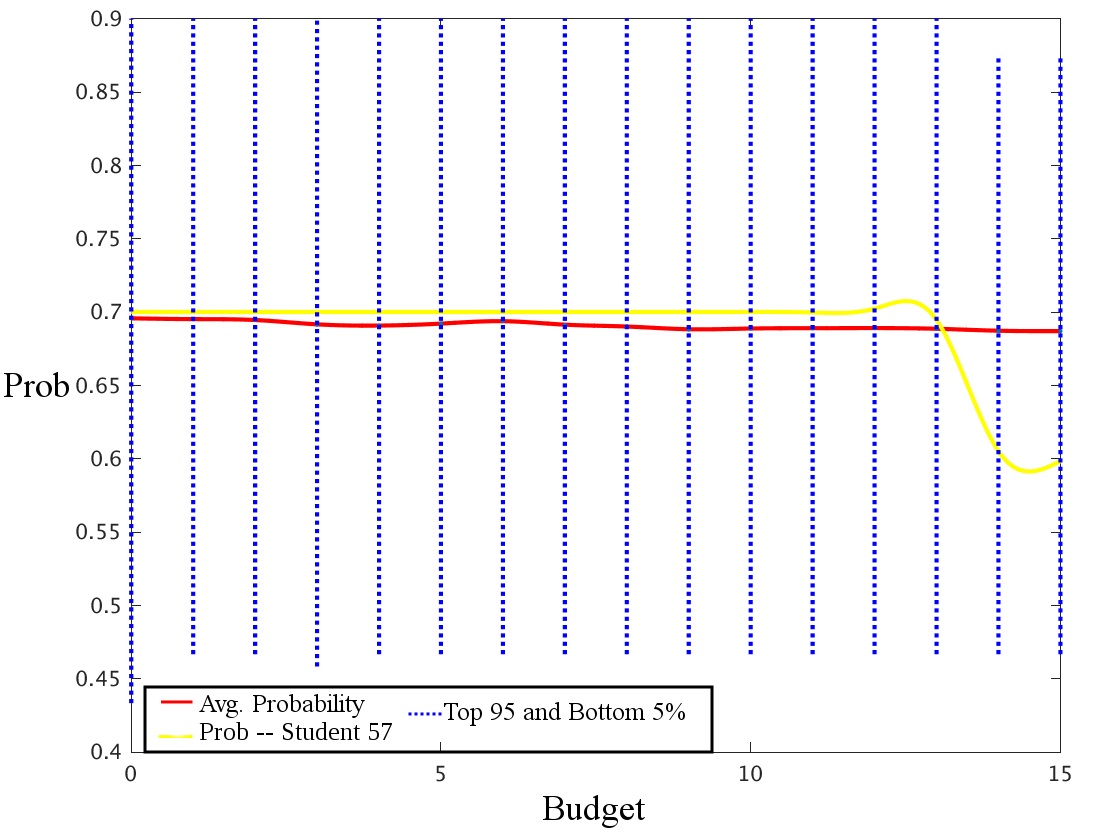}
			\caption{LVP-BI
				\label{fig:benchlvpbi_prob}}
		\end{subfigure}\par
		\begin{subfigure}[h]{\linewidth}
			\centering
			\includegraphics[width=\linewidth]{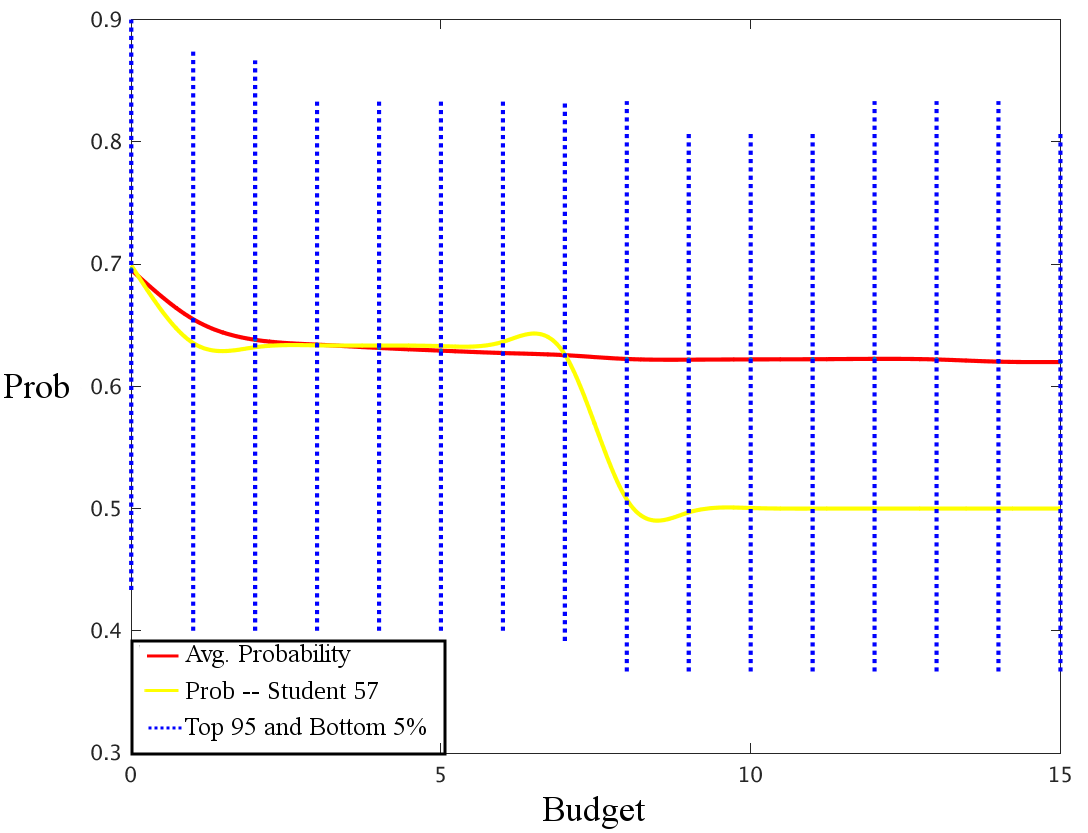}
			\caption{HC+LS
				\label{fig:phcbench_prob}}
		\end{subfigure}\par
\end{figure}

\begin{figure}[H]
			\begin{subfigure}[h]{\linewidth}
				\centering
				\includegraphics[width=\linewidth]{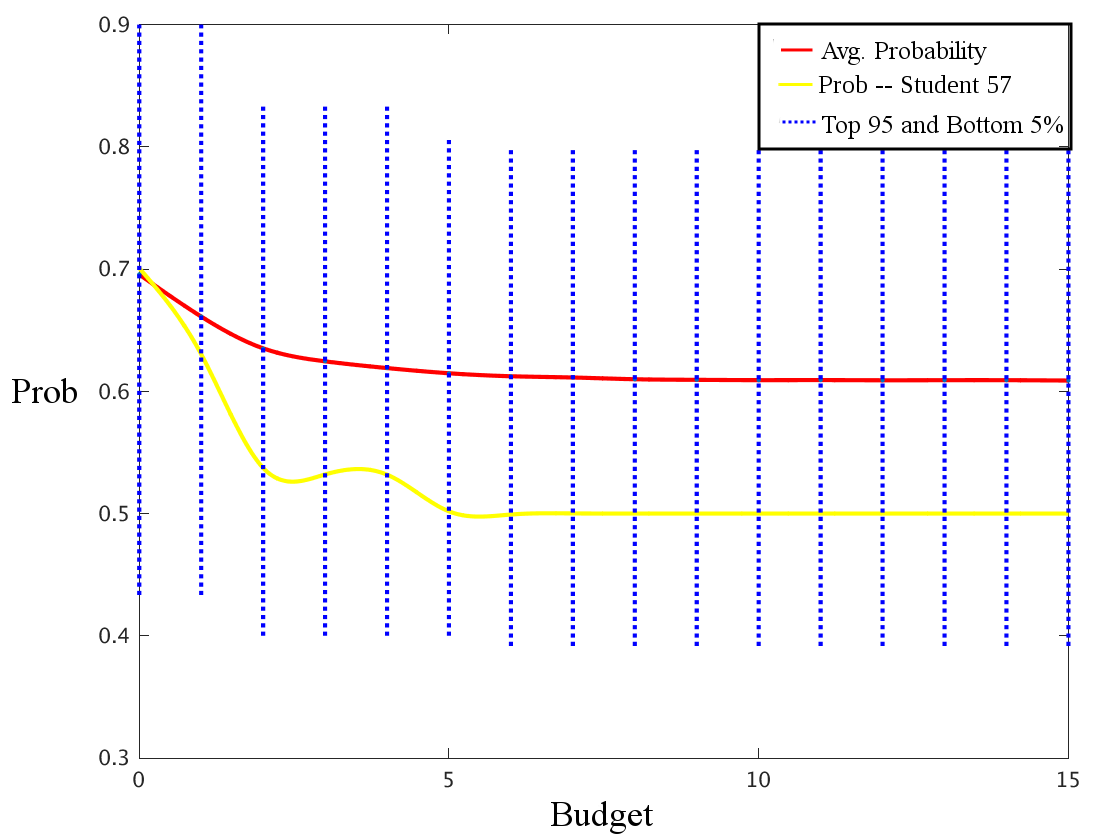}
				\caption*{(d) GA
					\label{fig:pgabench_prob}}
			\end{subfigure}\par
			\begin{subfigure}[]{\linewidth}
				\centering
				\includegraphics[width=\linewidth]{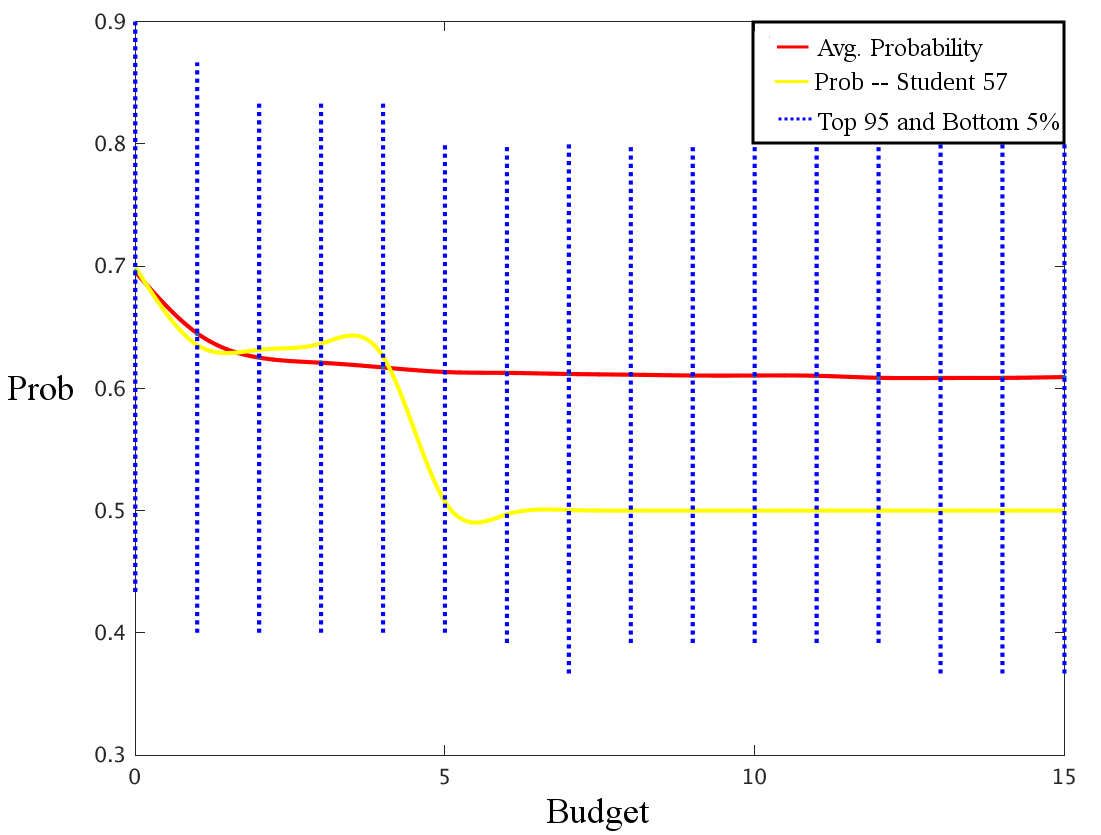}
				\caption*{(e) GA+LS
					\label{fig:pglbench_prob}}
			\end{subfigure}
			\caption{Probability vs budget for the five methods.}
			\label{fig:benchprop}
\noindent\makebox[\linewidth]{\rule{\linewidth}{0.4pt}}
\end{figure}

\begin{figure}[H]
	\centering
		\begin{subfigure}[h]{\linewidth}
			\centering
			\includegraphics[width=\linewidth]{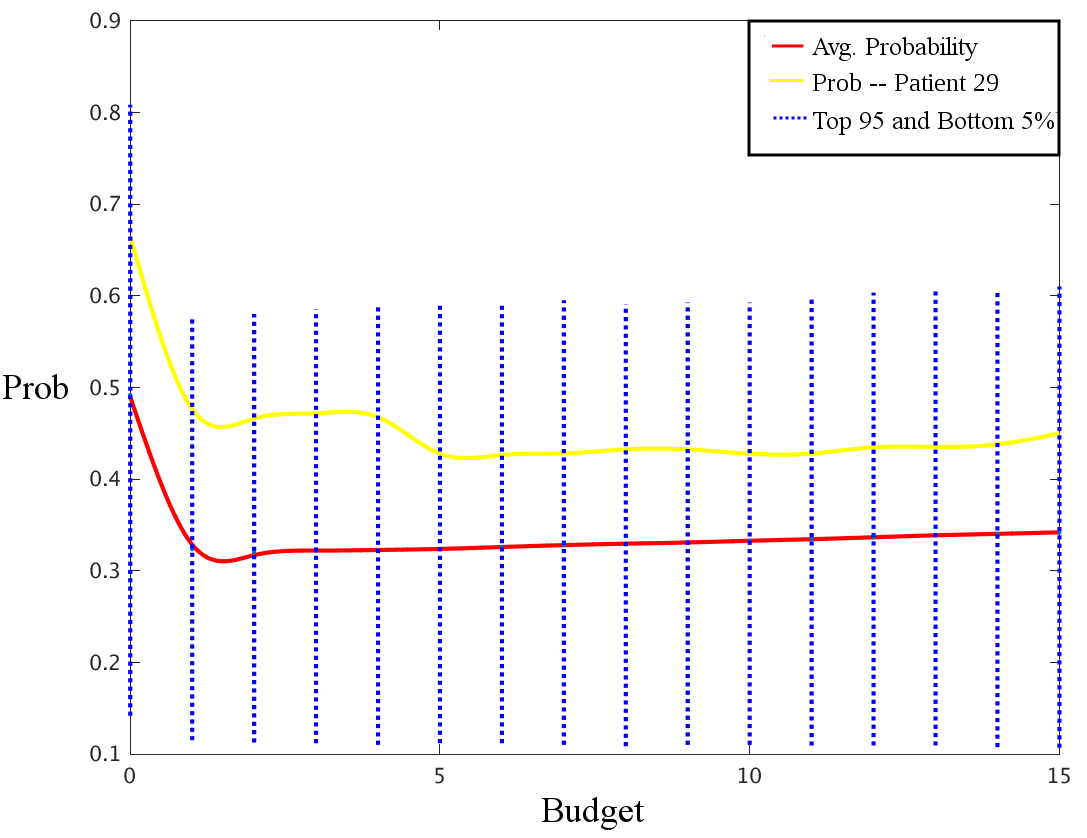}
			\caption{LVP-FI \label{fig:cvdlvpfi_prob}}
		\end{subfigure}\par
		\begin{subfigure}[h]{\linewidth}
			\centering
			\includegraphics[width=\linewidth]{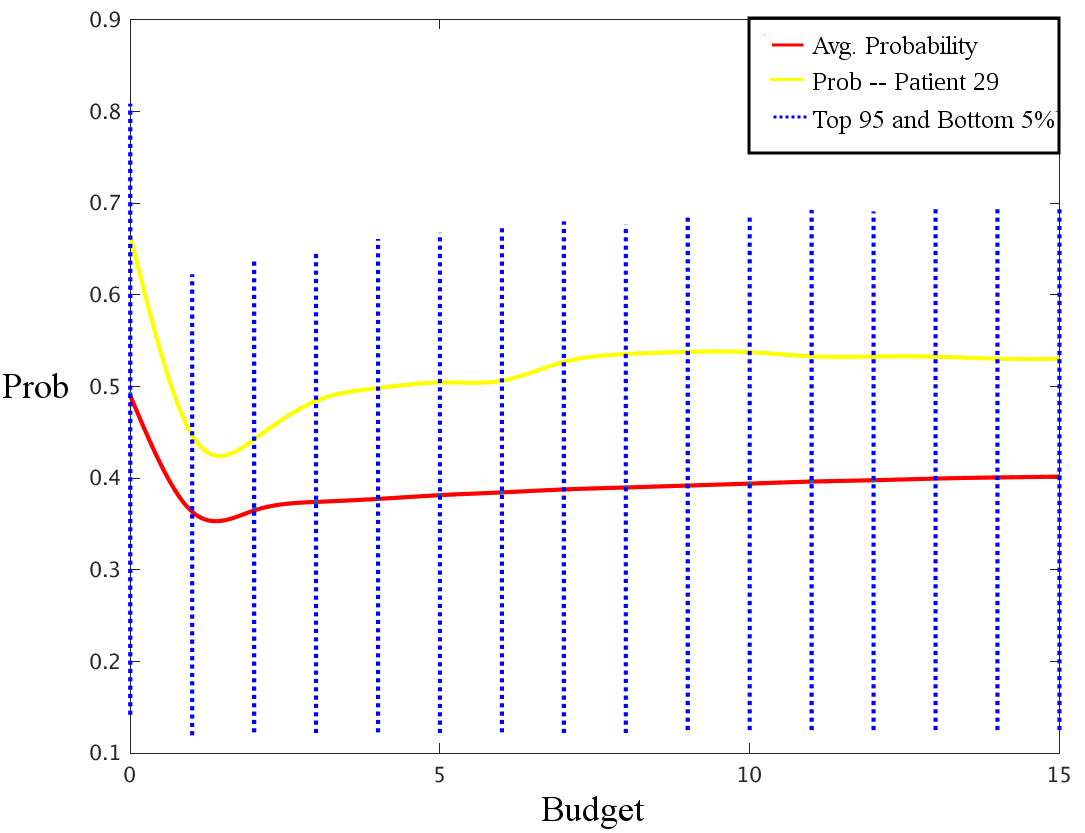}
			\caption{LVP-BI \label{fig:cvdlvpbi_prob}}
		\end{subfigure}
		\begin{subfigure}[h]{\linewidth}
			\centering
			\includegraphics[width=\linewidth]{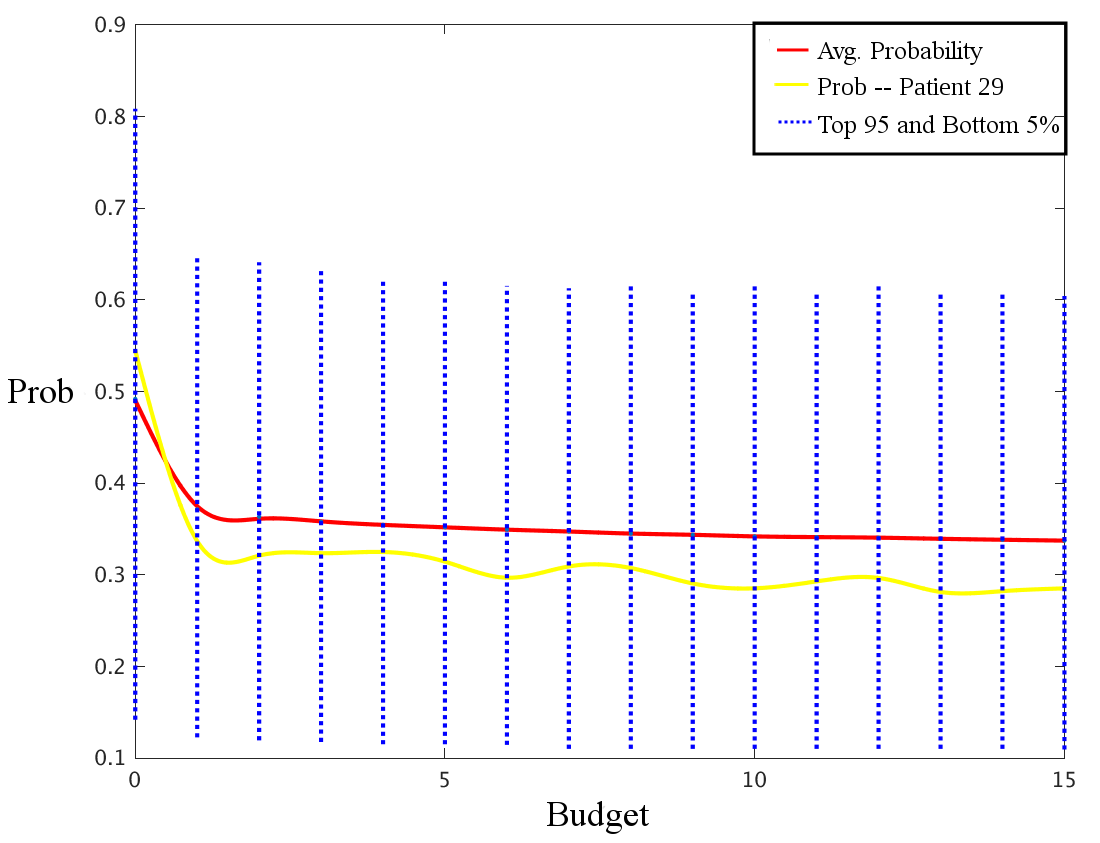}
			\caption{HC+LS \label{fig:cvdhc_prob}}
		\end{subfigure}\par
\end{figure}

\begin{figure}[H]
		\begin{subfigure}[H]{\linewidth}
			\centering
			\includegraphics[width=\linewidth]{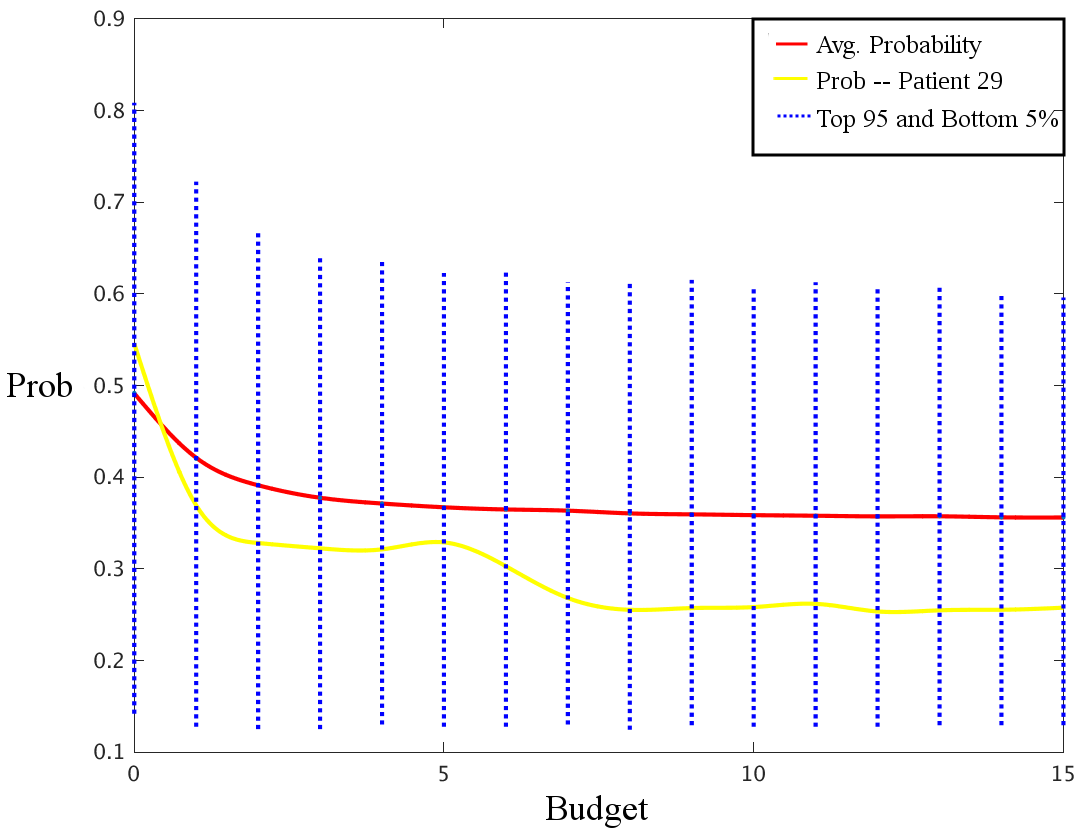}
			\caption*{(d) GA
				\label{fig:cvdga_prob}}
		\end{subfigure}\par
		\begin{subfigure}[H]{\linewidth}
			\centering
			\includegraphics[width=\linewidth]{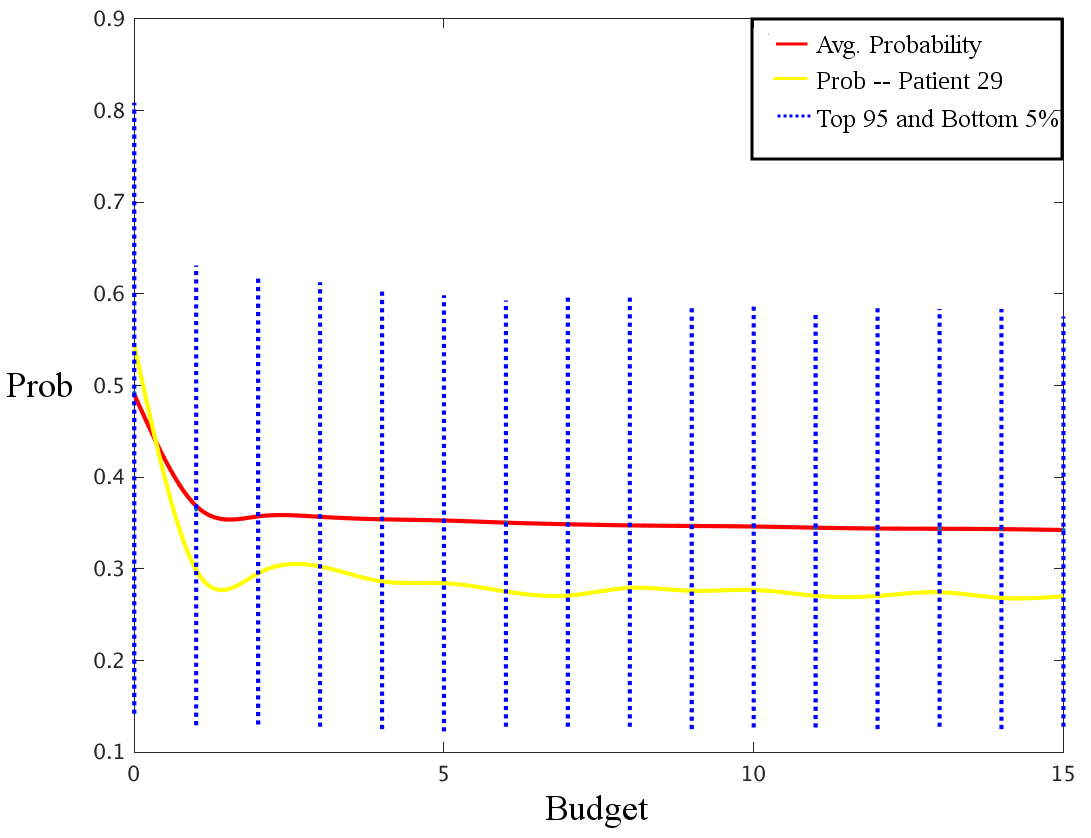}
			\caption*{(e) GA+LS
				\label{fig:cvdgal_prob}}
		\end{subfigure}	
		\caption{Probability vs budget for the five methods. \label{fig:cvdProp}}
\noindent\makebox[\linewidth]{\rule{\linewidth}{0.4pt}}
\end{figure}

\end{document}